\documentclass{article} %
\usepackage[preprint]{colm2025_conference}

\usepackage{microtype}
\usepackage{hyperref}
\usepackage{url}
\usepackage{booktabs}

\usepackage{lineno}

\usepackage{multirow}
\usepackage[normalem]{ulem}
\usepackage{listings}   %
\usepackage{graphicx}   %
\usepackage{amsmath}
\usepackage{fdsymbol}   %
\usepackage{xspace}
\usepackage{colortbl}
\usepackage{tcolorbox}
\usepackage{fontawesome5}

\newtcolorbox{shadedlisting}{
    colback=gray!10,
    boxrule=0pt,
    left=15pt,
    right=15pt,
    top=10pt,
    bottom=10pt
}

\newtcolorbox{qualitativeBox}{
  colback=gray!10,
  colframe=gray!20,
  boxrule=0.5pt,
  left=10pt,
  right=10pt,
  top=5pt,
  bottom=5pt,
  boxsep=5pt,
}

\newcommand{\gold}[0]{D_{\text{gold}}}
\newcommand{\synth}[0]{D_{\text{gen}}}

\newcommand{\fprobe}[0]{f_{\text{int}}}
\newcommand{\fdoc}[0]{f_{\text{ext}}}
\newcommand{\analysistool}{{\textsc{Epistemic Training}{}}}

\definecolor{forestgreen}{rgb}{0.13, 0.55, 0.13}

\newcommand{\arredit}[1]{{#1}}

\definecolor{darkblue}{rgb}{0, 0, 0.5}
\hypersetup{colorlinks=true, citecolor=darkblue, linkcolor=darkblue, urlcolor=darkblue}

\title{The Curious Case of Factuality Finetuning: \\Models' Internal Beliefs Can Improve Factuality}

\author{Benjamin Newman$^{\spadesuit}$ \quad Abhilasha Ravichander$^{\spadesuit}$ \textbf{Jaehun Jung}$^\spadesuit$ \quad \textbf{Rui Xin}$^\spadesuit$ \\ 
\textbf{Hamish Ivison}$^\spadesuit$ \quad \textbf{Yegor Kuznetsov}$^\spadesuit$ \quad \textbf{Pang Wei Koh}$^{\spadesuit}$ \quad \textbf{Yejin Choi}$^\diamondsuit$
\vspace{0.5em}
\\
$^\spadesuit$University of Washington \quad
$^\diamondsuit$Stanford University
\vspace{0.5em}
\\
\texttt{blnewman@cs.washington.edu}
}

\begin{document}

\ifcolmsubmission
\linenumbers
\fi

\maketitle

\begin{abstract}
 Language models are prone to \textit{hallucination}---generating text that is factually incorrect. Finetuning models on high-quality factual information can potentially reduce hallucination, but concerns remain; obtaining factual gold data can be expensive and training on correct but unfamiliar data may potentially lead to even more downstream hallucination.
 What data should practitioners finetune on to mitigate hallucinations in language models?
 In this work, we study the relationship between the factuality of finetuning data and the prevalence of hallucinations in long-form generation tasks. Counterintuitively, we find that finetuning on factual gold data is not as helpful as finetuning on model-generated data that models \textit{believe} to be factual. Next, we evaluate filtering strategies applied on both factual gold data and model-generated data, and find that finetuning on model-generated data that is filtered by models’ own internal judgments often leads to better overall factuality compared to other configurations: training on gold data filtered by models' judgments, training on gold data alone, or training on model-generated data that is supported by gold data. These factuality improvements transfer across three domains we study, suggesting that a models' own beliefs can provide a powerful signal for factuality.\footnote{\quad \faGithub{} \xspace{} \href{https://github.com/bnewm0609/epistemic-training}{\texttt{github.com/bnewm0609/epistemic-training}}}
\end{abstract}

\section{Introduction}

A significant obstacle limiting widespread use of language models (LMs) is model hallucination---models tend to generate text that is factually incorrect without any indication to readers \citep{wang2024factuality}.
This problem is particularly acute in long-form generation tasks, where generations can be partially correct \citep{Huang_Liu_Thirukovalluru_Cohan_Dhingra_2024}, and factual errors might accumulate during generation \citep{Zhang_Press_Merrill_Liu_Smith_2023}.
As a result, finetuning models to improve factuality has become a standard component of post-training recipes \citep{dubey2024llama, yang2023alignment, zhang2023rtuning}.
However, this can be challenging---recent work in short-form question answering has noted that finetuning on factually correct but \textit{unfamiliar} knowledge can actually hurt downstream factuality; accordingly, these works recommend finetuning on knowledge that is both factually correct and familiar \citep{Gekhman_Yona_Aharoni_Eyal_Feder_Reichart_Herzig_2024, Ghosal_Hashimoto_Raghunathan_2024}.

\begin{figure*}[th]
    \centering
    \includegraphics[width=\linewidth]{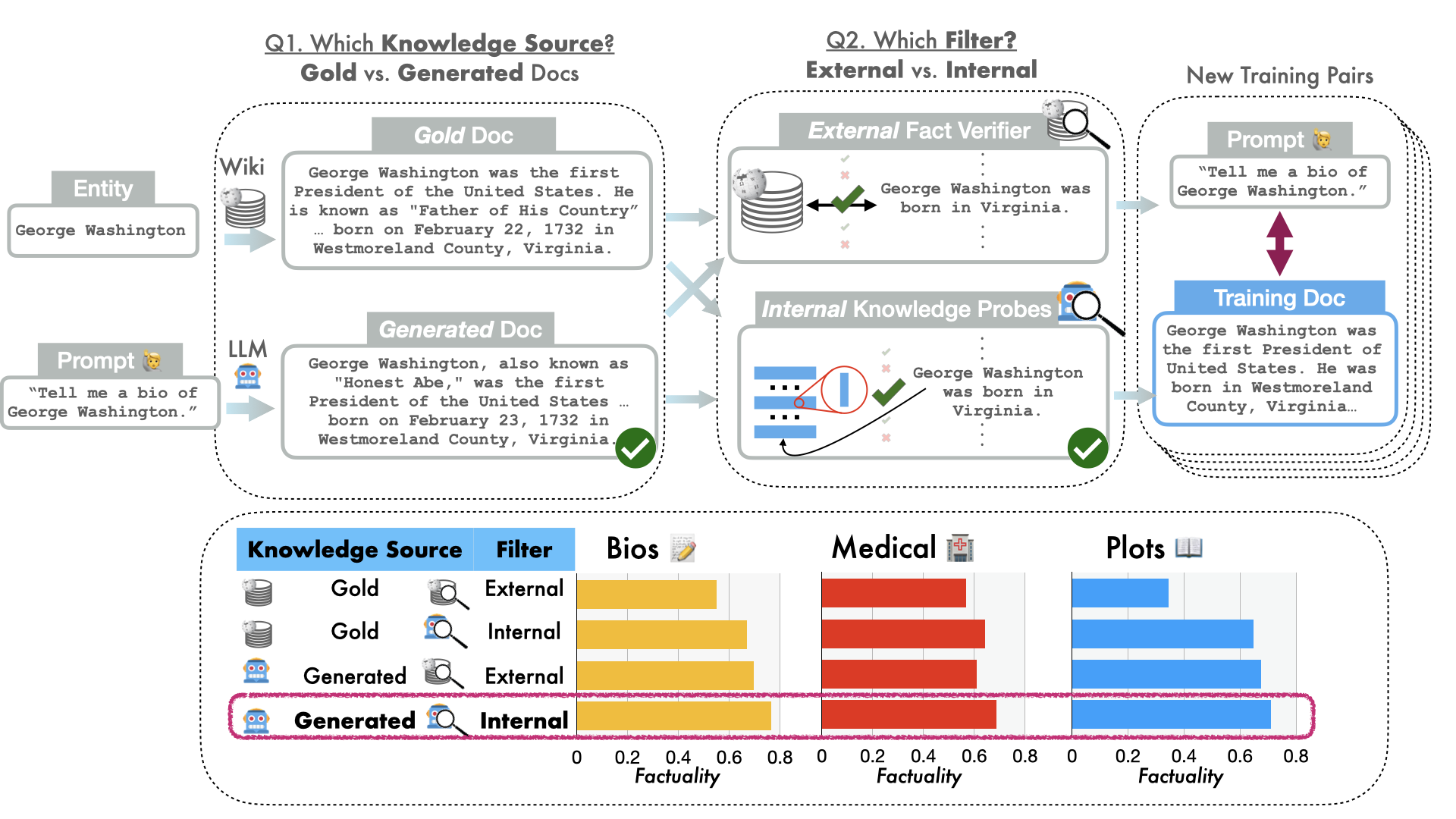}
    
    \caption{We manipulate two aspects of finetuning data: the \textit{knowledge source} (using the gold document directly or employing an LM to generate a document based on the entity discussed in the gold document) and \textit{filtering method} (entailment from external documents or probes on model hidden states) used for curating training data to improve finetuned model factuality.
    \analysistool{} (bolded)---training on generated data, filtered with internal knowledge probes (internal) outperforms both training on gold data as well as generated data filtered with entailment to gold documents (external). %
    }
    \label{fig:overview}
\end{figure*}

In this work, we find that model factuality improves even when finetuning on data that is \textit{less factual}---in the settings we study, factuality is often improved most when the finetuning data aligns with models' internal knowledge, even if the knowledge is unsupported by gold documents. %
We come to this conclusion by manipulating two attributes of finetuning data in controlled long-form generation experiments (outlined in Figure~\ref{fig:overview}).

First we consider the \textit{knowledge source} used to create the finetuning data, and we compare finetuning on gold data to training on the models' own generations. While the first is more factual, we find that finetuning on the latter leads to greater improvements in factuality when describing a held out set of entities.

Second, we consider how to \textit{filter} the training data. Long form generations may contain incorrect claims, so we filter claims from the generations before training on them. We compare filtering using external knowledge (only keeping information entailed by gold documents) to filtering based on internal knowledge (only keeping facts that a probe predicts are true from models' hidden states). We use this latter technique to represent what models \textit{believe} to be true, and it is motivated by prior work indicating that probes trained on model representations can predict whether models’ own generations are factual \citep{Azaria_Mitchell_2023, Liu_Casper_Hadfield-Menell_Andreas_2023, Joshi_Rando_Saparov_Kim_He_2023}.
Compared to training on data filtered with external knowledge, using internal knowledge filters often leads to a greater increase in factuality on descriptions of held-out entities, despite the training data being less factual.

Based on these observations, we introduce \analysistool{},
a procedure to post-train models using data generated by, and filtered using, the model's own internal knowledge---the best performing combination of knowledge source and filter.
We use \analysistool{} to analyze the dependence of models on the factuality of their training data and find that it increases factuality compared to training on gold data alone by up between 12--30 percentage points (Figure \ref{fig:overview} top-most versus bottom-most columns).

Finally, while \analysistool{} requires less gold data compared to other methods, it still requires a small amount for training the internal knowledge filter.
As such, we also investigate whether even this gold data is necessary by studying cross domain transfer---whether models finetuned with \analysistool{} on one domain can perform well on others.
In general, we find evidence that factuality does generalize across domains although using in-domain data when possible is better.

We summarize our contributions as follows:
\begin{enumerate}
    \item We extend the findings that models trained on gold data are less factual than models trained on generated data \citep{Gekhman_Yona_Aharoni_Eyal_Feder_Reichart_Herzig_2024, Ghosal_Hashimoto_Raghunathan_2024} to long-form generation settings.
    \item We find that filtering claims using probes on models' hidden states improves factuality over filtering using using entailment to external documents.
    \item We find that the factuality gains learned on one domain can transfer to improvements on others.
\end{enumerate}

Our analysis sheds light on how model practitioners should filter data for post-training, as well as provides an understanding of specific signals that models can leverage to improve their factual accuracy.

\section{Approach}
In order to investigate how the factuality of training data affects models' hallucination rates, we compare different methods for creating finetuning data using different knowledge sources and factuality filters.

Our data creation pipeline has three stages.
First, we choose a \textit{knowledge source}: selecting gold training documents or generating synthetic documents to create an initial training set. Second we choose a \textit{filtering strategy} to remove incorrect claims from those documents. Finally we merge the remaining claims together to construct a final training set.
We finetune models on this data and measure their hallucination rates.
See Figure~\ref{fig:overview} for an overview.

\subsection{Knowledge Source}
We compare training on two knowledge sources: gold documents, $\gold$, and model-generated documents, $\synth$.
Formally, we consider a set of entities $E$, and documents $\gold$ where each entity $e_k$ has an associated human-authored gold document, $d_k \in \gold$, that contains a large amount of factual information about $e_k$.
(Note that in our setting we have a separate $E$ and $\gold$ for training and evaluation.)

\paragraph{Generated Documents ($\synth$).} 
Our data for $\synth$ is generated as follows:
We place each $e_k$ in a prompt template and sample 10 generations per prompt, $y^i_k$, where $i$ is the generation index.\footnote{
We generate by sampling with a temperature of $0.7$, and a maximum of $1000$ new tokens.}
The prompts differ by dataset, but usually ask for information about some entity or fact.
For example, a prompt might be \textit{``Give me a biography of George Washington.''}.\footnote{See Appendix \ref{appendix:dataset-details} for the exact prompts templates.}
Taken together, these samples represent an initial source of generated training data, $\synth$. More formally, $\synth = \left\{y^i_k \mid i \in \{1, \ldots, 10\}, k \in \{1, \ldots |D|\}\right\}$.

\subsection{Filtering Strategy}

The next step in our pipeline is to identify and remove incorrect claims from the data sources.
To do this, we break each sample down into a sequence of atomic claims, and then we filter out the claims that are incorrect.
We compare two methods for filtering out incorrect claims: one using internal knowledge and one using external document knowledge.

\paragraph{Atomization.}
The first step of filtering is to \textit{atomize} each sample $z_k$ from $\gold$ and $\synth$ into a sequence of atomic claims $[a_k^1, a_k^2, ..., a_k^m]$ (as done in \citet{Min_Krishna_Lyu_Lewis_Yih_Koh_Iyyer_Zettlemoyer_Hajishirzi_2023}).
We perform atomization by first splitting the sample into sentences, and then passing each sentence to a model along with few-shot examples. \footnote{We use Mistral-7B-Instruct-v0.2 \citep{jiang2023mistral7b}, and see Appendix~\S\ref{appendix:prompt-atomize} for the prompt and Appendix~\S\ref{appendix:atomization-choice} for validation of the model.}

Next, we identify which atomic claims are correct.
We define a filter function $f(a_{k}^m)$ that outputs \texttt{supported} if $a_{k}^m$ should be considered ``correct'' and \texttt{unsupported} otherwise.
We compare two such $f$: one using external document knowledge $\fdoc$, and one using internal knowledge, $\fprobe$.

\paragraph{External Fact Verifier ($\fdoc$).}
This filter outputs \texttt{supported} if $a_{k}^m$ is entailed by the associated gold document, $d_k$.\footnote{For our purposes, we assume that the $d_k$ contains all relevant information about $e_k$. See Appendix~\ref{sec:limitations} for more discussion.}
To compute this entailment, we use a MiniCheck model~\citep{tang2024minicheck} (based on Flan-T5-large), as it is fast and accurate, and pass both the atomic fact and original gold document.\footnote{See Appendix \ref{appendix:entailment-comp} for validation of Minicheck's quality.}
\arredit{Following \citet{tang2024minicheck}}, we mark the atom as \texttt{supported} if the Minicheck model returns a higher than 50\% probability of claim being supported by the document and \texttt{unsupported} otherwise.
For $\gold$, we do not perform this filtering, as we assume that all claims from the gold document are correct.

\paragraph{Internal Knowledge Probes ($\fprobe$).}
Our second filter function is computed with \textit{internal knowledge probes}.
This filter works by encoding each claim from each document separately and applying a binary logistic regression classifier (the probe) to the encoding's hidden states to predict if the claim is \texttt{supported} or \texttt{unsupported}.
Following prior work and a pilot study, we find that applying these probes to the last token at a middle layer of the model leads to the best classification F1 \citep{CH-Wang_Durme_Eisner_Kedzie, Joshi_Rando_Saparov_Kim_He_2023}.
(See Appendix \S\ref{appendix:probe-f1} for more details.)
To train these classifiers, we use documents associated with a held out set of entities for each of our datasets as inputs, and silver labels generated by $\fdoc$ as training labels.
Each unique dataset, model, and knowledge source combination has its own probes.
While labeled data is used to train these probes, we still consider them to reflect internal knowledge because they are trained on a separate set of entities  from the training and evaluation entities.

\subsection{Dataset Construction}
Finally, we combine the \texttt{supported} claims that remain after filtering to reconstruct our training data.
To do this, we remove all atoms marked as \texttt{unsupported} by the filter, and merge the remaining atoms together using an LM. \footnote{Llama 3 8B Instruct \citep{dubey2024llama}; see Appendix~\S\ref{appendix:prompt-merge} and \S\ref{appendix:merging-choice} for the prompt and details.}
If there are no factual atoms for a given prompt, we replace the model output with a refusal message. Either \textit{`I'm sorry, I don't know much about [entity].'} or \textit{`I'm sorry, I'm unable to fulfill your request.'}, depending on the dataset.

\section{Experiments}

Our main experiments compare the effects of using $\gold$ and $\synth$ as the data sources and filtering those sources with $\fdoc$ and $\fprobe$.
In addition, we compare to a baseline with no additional finetuning.
We experiment with three models that have been studied for factuality by prior works (e.g. \cite{alnuhait2024factcheckmate, Tian_Mitchell_Yao_Manning_Finn_2023}): Mistral-7B-Instruct-v0.2 \citep{jiang2023mistral7b}, Llama-2 7B Chat \citep{touvron2023llama}, and Gemma 7B Instruction Tuned \citep{Mesnard2024GemmaOM}. %

\paragraph{Training.}
We finetune LMs on these datasets using LoRA~\citep{hu2021lora} to reduce computational cost.
We use a rank of 8 with $\alpha=16$ and dropout of 0.05.
We use a learning rate of $3\times10^{-4}$ and train for 500 steps with packed sequences and an effective batch size of 16 samples.\footnote{All values are found through hyperparameter search (See Appendix \ref{appendix:hyperparam-search}).}
Because we sample 10 generations per entity when creating synthetic data, we train on the gold data for ten times as many steps.

\paragraph{Evaluation.} We evaluate our models by generating five samples from a held-out set of prompts, and then atomizing the claims and determining how many are correct as judged by the MiniCheck model and the gold documents (i.e., we use the same approach as in the external knowledge filtering setting---$\fdoc$).
We report three metrics: the \textit{factuality}, \textit{detail}, and \textit{abstention rate} of model generations.

Factuality is the average percentage of supported atoms in model generations\footnote{This metric is similar to that of \textsc{FActScore} \citep{Min_Krishna_Lyu_Lewis_Yih_Koh_Iyyer_Zettlemoyer_Hajishirzi_2023} but uses open-weight models.}; Detail is the average number of atoms across non-abstaining model generation; and abstention rate is the percentage of generations where the model outputs a refusal message. For determining abstentions, we use the simple heuristic of the presence of a first person pronoun in the first sentence of the model generation (e.g. ``I'm sorry...'' or ``I don't know...''; this heuristic was validated by manual inspection).

\paragraph{Length Control.}
In pilot experiments, we found that the length of the model generations affected their factuality, an observation that has also been noted in prior work \citep{Min_Krishna_Lyu_Lewis_Yih_Koh_Iyyer_Zettlemoyer_Hajishirzi_2023, Zhang_Press_Merrill_Liu_Smith_2023}.
This presents a problem as $\fprobe$ tends to filter out more atoms than $\fdoc$, making it unclear if observed factuality differences were due to length or the semantic content of the training data.\footnote{See Appendix~\ref{appendix:factuality-vs-length} for an illustration.}
As such, we decided to control for length by performing additional filtering across our four conditions.
For each training sample, we constrain the number of atoms included in the training set to be $p$, the minimum number of \texttt{supported} atoms across the four conditions: ($\{\gold, \synth \} \times \{\fdoc, \fprobe\}$).

To do this, we rank each of the atoms and take the top $p$. For data filtered with $\fprobe$, we use the probability assigned by the probe to rank the supported atoms.
For $\synth$ filtered with $\fdoc$, we use the score assigned by the entailment function.
And for $\gold$ filtered with $\fdoc$, we use the first $p$ atoms in the document, as all of the facts from the document are \texttt{supported}.
For each sample, these $p$ atoms are merged together using Llama-3-8B as described above.

\paragraph{Datasets}
\label{sec:datasets}
For our experiments, we use three long-form generation datasets: biographies (bios), book and movie plots (plots), and medical term definitions (medical).\footnote{These datasets are derived from Wikipedia text, which is licensed under the CC BY-SA license. See Appendix \ref{appendix:dataset-details} for additional details and prompts.}

\textbf{Bios}: This dataset provides a list of names along with that person's  
Wikipedia page as a reference. 
We extract $\sim$9,000 entities from Wikidata that are instances of  ``human'' (\texttt{Q5}), and have an English Wikipedia page.
When computing labels using $\fdoc$ or for training $\fprobe$, we use the entire document; however as the Wikipedia pages for many of these entities are quite long, when constructing $\gold$, we use the opening section of the Wikipedia page as the gold document.

\textbf{Plots}: This dataset provides contains $\sim$3,000 media titles (books, movies, plays, etc.) along with their plot summaries extracted from Wikipedia by \citet{riedl-2017-wikiplots}. 

\textbf{Medical}~\citep{wiki_medical_term}: This dataset provides a set of $\sim$800 specific medical terms (\textit{e.g., Lymphogranuloma venereum}) along with  corresponding Wikipedia articles as references. 

\begin{table}
\centering
\small
\renewcommand*{\arraystretch}{1.2}
\begin{tabular}{@{}lccc@{}}
\toprule
\textbf{Dataset} & \textbf{\# Train} & \textbf{\# Probe Train} & \textbf{\# Test} \\ \midrule
Bios & 7207 & 250 & 932 \\
Plots  & 2257 & 151 & 301 \\
Medical & 500 & 199 & 100 \\
\bottomrule
\end{tabular}
\caption{Sizes and domains of the datasets we examine in this work. We report number of entities from each dataset used for each split.}
\label{tab:datasets_table}
\end{table}

For each dataset, we construct a random train-test split along with a small set of entities for training the probes (sizes in Table~\ref{tab:datasets_table}).

\section{Results}
\newcommand{\ccolor}{gray!25}
\begin{table*}[t]\centering
\resizebox{1.0\textwidth}{!}{
\setlength{\tabcolsep}{2pt}
\begin{tabular}{ccc|ccccccccc}
\toprule
\multicolumn{3}{c|}{\textbf{Data Construction Method}} & \multicolumn{3}{c}{\textbf{Mistral-7B-Instruct}} & \multicolumn{3}{c}{\textbf{Gemma-7B-it}} & \multicolumn{3}{c}{\textbf{Llama-2-7B-Chat}} \\ \cmidrule(lr){1-3} \cmidrule(lr){4-6} \cmidrule(lr){7-9} \cmidrule(lr){10-12}
 Data & KS & Filter & F ($\uparrow$) & D ($\uparrow$) & A ($\downarrow$) & F ($\uparrow$) & D ($\uparrow$) & A ($\downarrow$) & F ($\uparrow$) & D ($\uparrow$) & A ($\downarrow$) \\
\midrule
\multirow[c]{7}{*}{\textit{Bios}} & \textit{None} & \textit{None} & 29.6 & \bfseries 63.1 & 1.2 & 25.5 & \bfseries 47.0 & 0.0 & 31.6 & \bfseries 78.1 & 0.0 \\
\cmidrule{2-12}
 & \multirow[c]{2}{*}{\textit{Gold}} & 
 \textit{External} & 54.9$_{\text{(0.4)}}$ & 10.1$_{\text{(0.4)}}$ & 26.9$_{\text{(0.4)}}$ & 36.2$_{\text{(1.9)}}$ & 9.37$_{\text{(1.4)}}$ & 39.9$_{\text{(11.4)}}$ & 32.9$_{\text{(0.3)}}$ & 15.5$_{\text{(0.6)}}$ & 2.2$_{\text{(0.6)}}$ \\
 &  & \textit{Internal} & 66.9$_{\text{(0.8)}}$ & 10.6$_{\text{(0.1)}}$ & 32.2$_{\text{(2.3)}}$ & 42.6$_{\text{(4.9)}}$ & 10.5$_{\text{(0.6)}}$ & 36.2$_{\text{(11.3)}}$ & - & - & - \\
\cmidrule{2-12}
 & \multirow[c]{3}{*}{\textit{Gen}} & \textit{Random} & 53.4$_{\text{(0.9)}}$ & 12.5$_{\text{(0.7)}}$ & 28.5$_{\text{(6.0)}}$ & 43.3$_{\text{(0.3)}}$ & 7.80$_{\text{(0.2)}}$ & 47.9$_{\text{(1.6)}}$ & 42.2$_{\text{(1.8)}}$ & 15.5$_{\text{(0.8)}}$ & 5.6$_{\text{(0.8)}}$ \\
 &  & \textit{External} & 69.5$_{\text{(0.2)}}$ & 11.2$_{\text{(0.2)}}$ & 27.6$_{\text{(1.7)}}$ & 57.8$_{\text{(1.3)}}$ & 8.07$_{\text{(1.5)}}$ & 53.0$_{\text{(5.3)}}$ & 60.5$_{\text{(0.3)}}$ & 13.2$_{\text{(0.4)}}$ & 9.1$_{\text{(1.7)}}$ \\
 &  & \textit{Internal} & \bfseries \cellcolor{\ccolor} 76.6$_{\text{(0.9)}}$ & \cellcolor{\ccolor} 10.5$_{\text{(0.5)}}$ & \cellcolor{\ccolor} 29.3$_{\text{(0.8)}}$ & \bfseries \cellcolor{\ccolor} 62.1$_{\text{(2.1)}}$ & \cellcolor{\ccolor} 8.11$_{\text{(1.8)}}$ & \cellcolor{\ccolor} 49.4$_{\text{(13.6)}}$ & \bfseries \cellcolor{\ccolor} 65.9$_{\text{(0.6)}}$ & \cellcolor{\ccolor} 12.7$_{\text{(0.6)}}$ & \cellcolor{\ccolor} 11.9$_{\text{(2.3)}}$ \\
\cmidrule{1-12}
\multirow[c]{7}{*}{\textit{Plots}} & \textit{None} & \textit{None} & 13.6 & \bfseries 10.0 & 0.0 & 28.7 & \bfseries 35.7 & 0.0 & 47.4 & \bfseries 21.1 & 0.0 \\
\cmidrule{2-12}
 & \multirow[c]{3}{*}{\textit{Gold}} & 
\textit{External} & 34.4$_{\text{(0.6)}}$ & 6.26$_{\text{(0.2)}}$ & 29.5$_{\text{(1.8)}}$ & 21.4$_{\text{(1.0)}}$ & 8.44$_{\text{(1.1)}}$ & 16.5$_{\text{(9.9)}}$ & 25.8$_{\text{(0.9)}}$ & 9.81$_{\text{(0.6)}}$ & 6.5$_{\text{(2.2)}}$ \\
 &  & \textit{Internal} & 64.8$_{\text{(2.1)}}$ & 6.11$_{\text{(0.6)}}$ & 33.9$_{\text{(5.5)}}$ & 36.9$_{\text{(2.7)}}$ & 10.1$_{\text{(2.0)}}$ & 5.5$_{\text{(2.7)}}$ & 37.3$_{\text{(0.9)}}$ & 10.6$_{\text{(0.3)}}$ & 8.8$_{\text{(1.8)}}$ \\
\cmidrule{2-12}
 & \multirow[c]{3}{*}{\textit{Gen}} & \textit{Random} & 52.1$_{\text{(4.1)}}$ & 6.31$_{\text{(0.7)}}$ & 66.5$_{\text{(5.3)}}$ & 34.5$_{\text{(0.5)}}$ & 11.5$_{\text{(1.7)}}$ & 10.5$_{\text{(8.7)}}$ & 50.5$_{\text{(1.2)}}$ & 10.9$_{\text{(0.2)}}$ & 11.4$_{\text{(1.9)}}$ \\
 &  & \textit{External} & 67.5$_{\text{(0.9)}}$ & 6.37$_{\text{(1.3)}}$ & 26.5$_{\text{(3.7)}}$ & 50.9$_{\text{(1.1)}}$ & 8.99$_{\text{(0.5)}}$ & 5.2$_{\text{(3.9)}}$ & 59.1$_{\text{(0.4)}}$ & 8.84$_{\text{(0.2)}}$ & 6.2$_{\text{(0.8)}}$ \\
 &  & \textit{Internal} & \bfseries \cellcolor{\ccolor} 70.9$_{\text{(1.2)}}$ & \cellcolor{\ccolor} 7.72$_{\text{(0.8)}}$ & \cellcolor{\ccolor} 32.2$_{\text{(2.5)}}$ & \bfseries \cellcolor{\ccolor} 52.3$_{\text{(0.3)}}$ & \cellcolor{\ccolor} 9.90$_{\text{(1.2)}}$ & \cellcolor{\ccolor} 4.6$_{\text{(3.0)}}$ & \bfseries \cellcolor{\ccolor} 59.2$_{\text{(0.6)}}$ & \cellcolor{\ccolor} 8.77$_{\text{(0.3)}}$ & \cellcolor{\ccolor} 7.4$_{\text{(0.8)}}$ \\
\cmidrule{1-12}
\multirow[c]{7}{*}{\textit{Medical}} & \textit{None} & \textit{None} & 34.6 & \bfseries 66.4 & 0.0 & 43.6 & \bfseries 23.9 & 0.0 & 33.2 & \bfseries 68.3 & 0.0 \\
\cmidrule{2-12}
 & \multirow[c]{3}{*}{\textit{Gold}} & 
 \textit{External} & 56.8$_{\text{(2.2)}}$ & 9.03$_{\text{(0.2)}}$ & 10.5$_{\text{(2.2)}}$ & 44.1$_{\text{(7.5)}}$ & 9.23$_{\text{(1.8)}}$ & 10.6$_{\text{(11.4)}}$ & 48.9$_{\text{(1.9)}}$ & 12.0$_{\text{(0.1)}}$ & 16.3$_{\text{(13.7)}}$ \\
 &  & \textit{Internal} & 64.1$_{\text{(0.5)}}$ & 9.93$_{\text{(1.2)}}$ & 15.7$_{\text{(14.0)}}$ & 54.8$_{\text{(4.2)}}$ & 8.01$_{\text{(1.3)}}$ & 7.3$_{\text{(4.5)}}$ & 54.7$_{\text{(1.5)}}$ & 13.4$_{\text{(0.9)}}$ & 20.7$_{\text{(9.0)}}$ \\
\cmidrule{2-12}
 & \multirow[c]{3}{*}{\textit{Gen}} & \textit{Random} & 41.1$_{\text{(1.8)}}$ & 12.7$_{\text{(1.3)}}$ & 5.5$_{\text{(3.0)}}$ & 47.9$_{\text{(2.1)}}$ & 10.1$_{\text{(1.6)}}$ & 14.0$_{\text{(9.2)}}$ & 41.2$_{\text{(2.7)}}$ & 13.5$_{\text{(0.6)}}$ & 9.1$_{\text{(7.4)}}$ \\
 &  & \textit{External} & 60.9$_{\text{(1.5)}}$ & 9.91$_{\text{(0.5)}}$ & 3.3$_{\text{(2.3)}}$ & 54.6$_{\text{(2.7)}}$ & 8.88$_{\text{(1.4)}}$ & 1.7$_{\text{(1.3)}}$ & 57.9$_{\text{(2.7)}}$ & 10.0$_{\text{(0.9)}}$ & 8.3$_{\text{(3.8)}}$ \\
 &  & \textit{Internal} & \bfseries \cellcolor{\ccolor} 68.8$_{\text{(0.7)}}$ & \cellcolor{\ccolor} 9.25$_{\text{(0.9)}}$ & \cellcolor{\ccolor} 18.3$_{\text{(17.4)}}$ & \bfseries \cellcolor{\ccolor} 60.7$_{\text{(1.2)}}$ & \cellcolor{\ccolor} 7.54$_{\text{(1.9)}}$ & \cellcolor{\ccolor} 4.3$_{\text{(4.0)}}$ & \bfseries \cellcolor{\ccolor} 65.1$_{\text{(3.9)}}$ & \cellcolor{\ccolor} 9.23$_{\text{(0.5)}}$ & \cellcolor{\ccolor} 17.5$_{\text{(11.2)}}$ \\
\bottomrule
\end{tabular}

}

\caption{Main results from fine-tuning Mistral-7B-Instruct-v0.2, Llama-2-7B-Chat, and Gemma-7B-it. KS=Knowledge Source, F=Factuality, D=Detail, and A=Abstention Rate. Numbers in parentheses are standard deviations across three random seeds for all sampling done in the pipeline. Dashes are where the Bios internal knowledge filter for Llama-2-7B-Chat predicts all claims are \texttt{unsupported}. 
\textbf{Using Generated with $\fprobe$ consistently leads to the highest factuality.} Factuality is percentage of correct claims, detail is average number of claims in non-abstaining responses, and abstentions is the percentage of abstaining responses.
Shaded rows represent \analysistool{}.
}

\label{tab:main_results_table}
\end{table*}

\begin{table}[ht]
\resizebox{1.0\textwidth}{!}{
    \centering
    \small
    \setlength{\tabcolsep}{2pt}
    \begin{tabular}{lccc|ccc|cc|ccc|ccc|ccc}
    \toprule
     & \multicolumn{6}{c}{\textit{Plots}} & \multicolumn{5}{c}{\textit{Bios}} & \multicolumn{6}{c}{\textit{Medical}} \\
    \cmidrule(r){2-7} \cmidrule(r){8-12} \cmidrule(r){13-18}
     & \multicolumn{3}{c}{\textit{Gold}} & \multicolumn{3}{c}{\textit{Generated}} & \multicolumn{2}{c}{\textit{Gold}} & \multicolumn{3}{c}{\textit{Generated}} & \multicolumn{3}{c}{\textit{Gold}} & \multicolumn{3}{c}{\textit{Generated}} \\
    \cmidrule(r){2-4} \cmidrule(r){5-7} \cmidrule(r){8-9} \cmidrule(r){10-12} \cmidrule(r){13-15} \cmidrule(r){16-18}
     & \textit{Mistral} & \textit{Llama-2} & \textit{Gemma} & \textit{Mistral} & \textit{Llama-2} & \textit{Gemma} & \textit{Mistral} & \textit{Gemma} & \textit{Mistral} & \textit{Llama-2} & \textit{Gemma} & \textit{Mistral} & \textit{Llama-2} & \textit{Gemma} & \textit{Mistral} & \textit{Llama-2} & \textit{Gemma} \\
    \midrule
    Q1 & 0.309 & \bfseries 0.114 & 0.166 & 0.055 & -0.012 & 0.016 & \bfseries 0.260 & \bfseries 0.090 & \bfseries 0.190 & 0.116 & 0.085 & \bfseries 0.104 & 0.033 & 0.072 & 0.104 & 0.075 & \bfseries 0.072 \\
    Q2 & \bfseries 0.358 & 0.101 & 0.117 & \bfseries 0.087 & \bfseries 0.034 & \bfseries 0.039 & 0.250 & 0.072 & 0.189 & \bfseries 0.127 & \bfseries 0.141 & 0.043 & 0.039 & \bfseries 0.097 & 0.072 & \bfseries 0.108 & 0.065 \\
    Q3 & 0.271 & 0.110 & 0.146 & 0.034 & -0.015 & 0.028 & 0.179 & 0.063 & 0.169 & 0.079 & 0.072 & 0.057 & 0.056 & 0.087 & 0.057 & 0.080 & 0.025 \\
    Q4 & 0.299 & 0.112 & \bfseries 0.181 & 0.034 & -0.007 & 0.006 & 0.092 & 0.089 & 0.065 & 0.026 & 0.055 & 0.063 & \bfseries 0.067 & 0.077 & \bfseries 0.109 & 0.085 & 0.059 \\
    \bottomrule
    \end{tabular}
}
    \caption{The average difference in factuality between generations from models trained with data filtered with $\fprobe$ and $\fdoc$. Q1 is the bottom quartile, Q4 is the top quartile. \textbf{Generally, terms in the bottom quartiles lead to most of the improvements.}}
    \label{table:factuality-vs-frequency}
\end{table}

The results of the comparisons between training on familiar and factual knowledge are in Table~\ref{tab:main_results_table}.
We report the \textit{factuality} (\% of atomized claims can be verified by the ground-truth answer), \textit{detail} (average \# of atoms in non-abstaining generations), and the \textit{abstention rate} (the proportion of model refusals) of responses from the model.
We find overall that \analysistool{}---training on $\synth$ filtered with $\fprobe$---achieves the best factuality across all models and datasets.
We comment on a number of specific comparisons below.

\paragraph{Model-generated data is better than gold data}

Across the different knowledge sources, we find that training on generated data tends to improve factuality compared to training on gold data by 4--33 percentage points.
We hypothesize that we observe this difference for similar reasons described by \citep{Gekhman_Yona_Aharoni_Eyal_Feder_Reichart_Herzig_2024, Ghosal_Hashimoto_Raghunathan_2024}.
The gold data likely includes entities that are less familiar to the model, which may in turn encourage the model to make claims about unfamiliar entities, leading to more hallucinations.
Perhaps surprisingly, the presence of these less familiar claims in the gold data hurts performance more than the presence of incorrect claims in the generated data.

To check if less familiar claims in the gold data hurt model performance, we study the relationship between the frequencies of entities in pretraining corpora and model factuality. 
We do not have access to the pretraining corpora of the models we test, so we use the Infini-gram API to obtain the frequency of evaluation entities from RedPajama as a proxy \citep{Liu2024InfiniGram, weber2024redpajama}.
We compare these frequencies to the difference between the factuality of models trained with data filtered using $\fprobe$ and $\fdoc$ on the same entities.
We expect these differences to be higher for the rare entities than for more frequent entities, indicating that the rarer entities benefit most from filtering with $\fprobe$.
We group the entities in quartiles by frequencies and compute the mean of the factualities in each quartile, visible in Table \ref{table:factuality-vs-frequency}. (For histograms of entity frequencies per dataset, see Appendix \ref{appendix:frequency}).
In general, the greatest benefits come in the bottom quartiles.%
These observations corroborate the observations of \citep{Ghosal_Hashimoto_Raghunathan_2024} in the different setting of long-form generation.

We also investigate if our $\fprobe$ can also help identify and filter these unfamiliar claims from the gold data.
We find that using these probes to filter training generations leads to improvements in factuality---6--30 percentage point increase. 
The exception is Llama-2-Chat in the Bios dataset, where the probe predicts all of the gold claims are \texttt{unsupported}.
Taken together, these results indicate that in general, these probes can help identify and filter such claims from gold data.

\paragraph{Filtering with internal knowledge is better than external documents}
Across the datasets and models, we observe that filtering using internal knowledge ($\fprobe$) tends to provide an improvement (of up to 8 points)
over filtering using entailment to external documents ($\fdoc$) for generated data ($\synth$), despite the data generated using $\fdoc$ being more factual ( Appendix~\ref{appendix:train-data-factuality}).
We can see that within each dataset, the detail is very similar, indicating that our length control was successful, however the detail is much lower than the baseline with no training.

To investigate why $\fprobe$ tends to perform better, we conduct a qualitative analysis of a set of claims where $\fprobe$ and $\fdoc$ disagree.
For our analysis, we randomly select  $\sim$100 claims from the biography dataset.
In general, we find that while all claims tend to be similar, claims predicted \texttt{supported} by only $\fprobe$ tend to be less detailed than ones predicted \texttt{supported} by only $\fdoc$.
(See examples in Table \ref{tab:qual_analysis_detail}).\footnote{Note that not all examples that $\fdoc$ predicts are \texttt{unsupported} are actually false---sometimes there are errors with entailment function or the data source may be incomplete (See \S~\ref{sec:limitations}).}
We quantify these observations by using the number of tokens in a claim as a proxy for its detail.
We find that the $\fdoc$-only \texttt{supported} claims have significantly more tokens than $\fprobe$-only \texttt{supported} claims as measured by a paired t-test for most cases (See Appendix~\ref{appendix:internal-vs-external-filtering}).

\paragraph{Does factuality training generalize across domains?}
\begin{table*}[t]\centering
\resizebox{.95\textwidth}{!}{
\begin{tabular}{ccc|ccccccccc}
\toprule
\multicolumn{3}{c}{\textbf{Experimental Setting}} & \multicolumn{3}{c}{\textbf{Mistral-7B-Instruct}} & \multicolumn{3}{c}{\textbf{Llama-2-7B-Chat}} \\ \cmidrule(lr){1-3} \cmidrule(lr){4-6} \cmidrule(lr){7-9}
 Test & Train & Filter & F ($\uparrow$) & D ($\uparrow$) & A ($\downarrow$) & F ($\uparrow$) & D ($\uparrow$) & A ($\downarrow$) \\
\midrule
\multirow[c]{5}{*}{\textit{Bios}} & \textit{None} & \textit{None} & 29.6 & \textbf{63.1} & 1.2 & 31.6 & \textbf{78.1} & 0.0 \\
\cmidrule{2-9}
 & \multirow[c]{2}{*}{\textit{Plots}} & \textit{External} & 67.9$_{\text{(2.7)}}$ & 13.4$_{\text{(1.9)}}$ & 50.1$_{\text{(12.4)}}$ & 52.6$_{\text{(0.6)}}$ & 11.7$_{\text{(1.4)}}$ & 34.3$_{\text{(3.8)}}$ \\
 &  & \textit{Internal} & \textbf{68.6$_{\text{(1.3)}}$} & 13.7$_{\text{(1.5)}}$ & 44.5$_{\text{(2.3)}}$ & \textbf{55.2$_{\text{(1.3)}}$} & 10.9$_{\text{(1.2)}}$ & 44.6$_{\text{(7.7)}}$ \\
\cmidrule{2-9}
 & \multirow[c]{2}{*}{\textit{Medical}} & \textit{External} & 50.5$_{\text{(2.4)}}$ & 9.63$_{\text{(1.5)}}$ & 5.1$_{\text{(2.4)}}$ & 46.6$_{\text{(1.9)}}$ & 8.92$_{\text{(0.6)}}$ & 35.9$_{\text{(7.2)}}$ \\
 &  & \textit{Internal} & 56.8$_{\text{(1.3)}}$ & 9.26$_{\text{(0.6)}}$ & 3.1$_{\text{(1.4)}}$ & 49.2$_{\text{(1.7)}}$ & 9.46$_{\text{(0.7)}}$ & 32.4$_{\text{(4.7)}}$ \\
\cmidrule{1-9}
\multirow[c]{5}{*}{\textit{Plots}} & \textit{None} & \textit{None} & 13.6 & \textbf{10.0} & 0.0 & 47.4 & \textbf{21.1} & 0.0 \\
\cmidrule{2-9}
 & \multirow[c]{2}{*}{\textit{Bios}} & \textit{External} & 58.8$_{\text{(1.9)}}$ & 9.96$_{\text{(1.1)}}$ & 3.3$_{\text{(1.3)}}$ & 51.9$_{\text{(0.8)}}$ & 14.4$_{\text{(0.8)}}$ & 0.4$_{\text{(0.3)}}$ \\
 &  & \textit{Internal} & \textbf{59.7$_{\text{(0.7)}}$} & \textbf{10.0$_{\text{(0.4)}}$} & 2.5$_{\text{(1.2)}}$ & \textbf{53.2$_{\text{(0.7)}}$} & 12.8$_{\text{(1.2)}}$ & 0.5$_{\text{(0.2)}}$ \\
\cmidrule{2-9}
 & \multirow[c]{2}{*}{\textit{Medical}} & \textit{External} & 47.9$_{\text{(2.1)}}$ & 8.92$_{\text{(0.2)}}$ & 0.1$_{\text{(0.1)}}$ & 38.2$_{\text{(3.5)}}$ & 7.65$_{\text{(0.7)}}$ & 36.2$_{\text{(8.1)}}$ \\
 &  & \textit{Internal} & 54.1$_{\text{(2.9)}}$ & 10.3$_{\text{(1.2)}}$ & 0.7$_{\text{(0.2)}}$ & 46.4$_{\text{(6.1)}}$ & 9.13$_{\text{(1.9)}}$ & 14.7$_{\text{(5.1)}}$ \\
\cmidrule{1-9}
\multirow[c]{5}{*}{\textit{Medical}} & \textit{None} & \textit{None} & 34.6 & \textbf{66.4} & 0.0 & 33.2 & \textbf{68.3} & 0.0 \\
\cmidrule{2-9}
 & \multirow[c]{2}{*}{\textit{Bios}} & \textit{External} & 52.9$_{\text{(0.3)}}$ & 21.0$_{\text{(0.5)}}$ & 0.1$_{\text{(0.1)}}$ & 49.1$_{\text{(0.9)}}$ & 26.3$_{\text{(1.2)}}$ & 0.0$_{\text{(0.0)}}$ \\
 &  & \textit{Internal} & \textbf{53.6$_{\text{(0.7)}}$} & 23.9$_{\text{(0.8)}}$ & 0.0$_{\text{(0.0)}}$ & \textbf{49.2$_{\text{(0.4)}}$} & 28.2$_{\text{(0.8)}}$ & 0.0$_{\text{(0.0)}}$ \\
\cmidrule{2-9}
 & \multirow[c]{2}{*}{\textit{Plots}} & \textit{External} & 50.6$_{\text{(0.1)}}$ & 17.7$_{\text{(3.0)}}$ & 1.6$_{\text{(2.1)}}$ & 47.1$_{\text{(0.7)}}$ & 21.6$_{\text{(0.6)}}$ & 0.5$_{\text{(0.1)}}$ \\
 &  & \textit{Internal} & 52.0$_{\text{(0.8)}}$ & 18.8$_{\text{(1.7)}}$ & 0.7$_{\text{(0.8)}}$ & 45.2$_{\text{(1.3)}}$ & 22.9$_{\text{(0.5)}}$ & 0.4$_{\text{(0.2)}}$ \\
\bottomrule
\end{tabular}
}

\caption{Cross-domain results from finetuning Mistral-7B-Instruct-v0.2 and Llama-2-7B-Chat on data from one domain and testing on others. F=Factuality, D=Detail, and A=Abstention Rate. Results are averaged across three seeds with standard deviations in parenthesis. Evaluation metrics are defined in Table \ref{tab:main_results_table}. \textbf{Models trained to be more factual on one domain also are more factual on other domains.} }
\label{tab:xdomain_results_table}
\end{table*}

Even though \analysistool{} performs well using fewer gold documents than other methods we try, it still requires a small set of human-authored texts for training the internal knowledge probes.
However, if it were possible to use \analysistool{} to train a model to be more factual in one domain, and generate more factual generations in a different domain, it would reduce the cost of scaling this method in the future.  
But should we expect cross domain generalization?
Prior work has motivated using probes for identifying hallucinations by claiming that they may pick up on some general signal of whether the model knows whether a claim is true or not \citep{slobodkin-etal-2023-curious, orgad2025llms}, while others argue that this generalization depends on the task \citep{orgad2025llms}.
If the former hypothesis is true, we might expect the signal to be conserved across domains, so a model trained to be more factual in one domain will also be more factual on others. 
To investigate this question, we use the Mistral-7B-Instruct and Llama-2-7B-Chat models trained on the synthetic data conditions on one dataset and run inference on the other two datasets.
The results are in Table \ref{tab:xdomain_results_table}.

Overall, we see that both filtering conditions increase factuality across domains compared to our baselines.
We also see a consistent, though small improvement of using models trained with \analysistool{} over using external fact verifiers.
Comparing across domains, we also observe that models trained on biographies tend to perform better on plots than medical and vice versa, though we are unable to make strong claims comparing across domains as we don't control for training sample length across domains.
In general, we see evidence that factuality does generalize across domain although using in-domain data when possible is better.

\paragraph{Does factuality improve only from removing facts?} While almost all the finetuning approaches improve factuality compared to the no-finetuning baseline, this improvement comes at a cost to the amount of detail and an increase in the abstention rate.
Detail is reduced in our finetuning settings because removing unsupported atoms from training data encourages models to generate shorter strings.

In our experiments, \textbf{we ensure that the factuality gains we see are actually due to removing incorrect atoms rather than just from reducing the amount of information the model generates.}
To test, we compare our filters to a random filter, which randomly selects $p$ atoms to keep without considering their factuality.%
\emph{We find that models trained with randomly filtered atoms are more factual than baselines but less factual than our training methods}, indicating that length is not the only driver of the observed factuality differences (See Table \ref{tab:main_results_table}).

The fact that factuality training always reduces the number of atoms generated by our models highlights a \textit{precision-recall trade-off} around model factuality: we can increase our model factuality by more aggressively removing atoms, but risk producing less informative generations.
A model that refuses to answer any question could be considered `perfectly factual', since it does not say anything incorrect---but such a model would not be useful.
In this work, we measure the average number of atoms as a way to highlight that our models are still producing non-trivial amounts of information,\footnote{We also provide qualitative analysis of the information content of generations in Appendix \ref{appendix:detail_analysis}} but investigating additional ways to closely control this trade-off or ways to add new knowledge without encouraging hallucinations are interesting future directions.

\section{Discussion}

\paragraph{Effect of knowledge source.}
We hypothesize that the differences between gold data and model-generated data we observe are due to claims in the gold data likely being less familiar to the model.
Other factors could contribute as well, such as differences in data quality. If the gold data is for some reason of lower quality, it could harm finetuned models.
We do not observe such quality differences, likely due to using the same filtering procedure and merging prompt for both sources. (See examples of gold and generated data from each domain in Appendix~\ref{appendix:sample-data}.)
Apart from quality, we do observe diversity differences; model-generated training data is slightly less diverse (Appendix~\ref{appendix:diversity}).
However, as the differences are small, it is unlikely that they are the driver of downstream factuality differences.

\paragraph{Limitations of Self-Training.} 
Our analysis reveals that using internal knowledge improves factuality; however, this success requires that the evaluated models possess knowledge of our test entities.
If the models lacked any knowledge about a particular domain, using our method might reduce hallucinations (i.e. by encouraging the model to abstain for all the outputs in the domain), but would not produce useful outputs.
Additionally, there might be concerns that training on self-generated data could lead to the propagation of societal biases.
For example in our biographies case, models might have less knowledge about people in certain gender or racial categories, and might disproportionately learn to abstain from generating biographies when prompted with names from those categories.
Both of these cases, teaching models new factual information and mitigating the expression of societal biases in self-training scenarios, present important directions for future work.

\section{Related Work}

\paragraph{Enhancing Model Factuality.} Prior work has proposed enhancing LM factuality by augmenting models with specialized decoding algorithms \citep{lee-2022-factuality, li2023inferencetime, varshney2023stitch}, incorporating relevant context \citep{Mishra_Asai_Balachandran_Wang_Neubig_Tsvetkov_Hajishirzi_2024, shi-etal-2024-trusting}, and optimizing prompts \citep{Jones_Palangi_Simões_Chandrasekaran_Mukherjee_Mitra_Awadallah_Kamar_2023, yang2023alignment}. 
However, mitigating hallucinations through post-training is particularly important for efficiency reasons. Alternatives such as inference-time techniques and post-editing methods increase the cost of each inference call, whereas post-trained models fit into current efficient inference approaches \citep{dubey2024llama, Kwon2023EfficientMM}.
Works in this area directly optimize models for factuality, either by curating externally grounded data for supervision \citep{huang-etal-2024-training, training-lms-on-kg} or modifying the alignment objective for factuality \citep{Tian_Mitchell_Yao_Manning_Finn_2023, yang2023alignment, Kang_Wallace_Tomlin_Kumar_Levine_2024, lin2024flame}.
In our work, we consider how models’ internal representations can be leveraged for curating post-training data.

\paragraph{Self-knowledge in Language Models.} Measuring models' self-knowledge---to what extent they know what they know---has been an active area research.
 Common methods include prompting models to verbalize uncertainty over their output \citep{Kadavath_Conerly_Askell_Henighan_Drain_Perez_Schiefer_Hatfield-Dodds_DasSarma_Tran-Johnson_et_al_2022, Xiong_Hu_Lu_Li_Fu_He_Hooi_2023, Tian_Mitchell_Zhou_Sharma_Rafailov_Yao_Finn_Manning_2023}, and training probes on model hidden states as done in our analysis \citep{Azaria_Mitchell_2023, Joshi_Rando_Saparov_Kim_He_2023, CH-Wang_Durme_Eisner_Kedzie, orgad2025llms, alnuhait2024factcheckmate}.
 Models' internal representation may identify truthful knowledge, even when the model outputs factually incorrect generations \citep{Liu_Casper_Hadfield-Menell_Andreas_2023, orgad2025llms}.
Based on findings from these works, we use models' internal representations to curate post-training data.

\paragraph{Data Familiarity and Factuality.} Finally, a growing number of works focused around short-form question-answering tasks reveal that naively training LMs on factual data may not reduce hallucinations---especially if the training data includes less familiar entities \citep{Mallen_Asai_Zhong_Das_Khashabi_Hajishirzi_2023a} that are poorly represented in the model's pretrained knowledge \citep{Gekhman_Yona_Aharoni_Eyal_Feder_Reichart_Herzig_2024, Ghosal_Hashimoto_Raghunathan_2024}.
Our results point to similar conclusions in text-generation tasks that require long-form responses, and suggest that by training on more familiar data, models can learn to generate facts that the model is aware of, rather than fabricating knowledge about what the model is unfamiliar with.
Moreover, unlike other works whose definitions of familiarity rely on direct search over a training corpus \citep{Mallen_Asai_Zhong_Das_Khashabi_Hajishirzi_2023a, carlini2023quantifyingmemorizationneurallanguage} or reward modeling \cite{Kang_Wallace_Tomlin_Kumar_Levine_2024}, our method grounds our equivalent to familiarity in models' internal representations. 

\section{Conclusion}

In this work, we study to what extent factual gold data is necessary when finetuning models to hallucinate less in long-form generation tasks.
Counterintuitively, we find that models hallucinate least when they \textit{believe} that the claims in their training data are true, a claim we verify by varying the \textit{knowledge source} used to create the training data (comparing model-generated data to gold data) as well as the \textit{filtering method} used (comparing filtering based on models' internal representations to entailment to gold documents).
These observations illustrate how to use minimal gold data to finetune models whose low hallucination rates can generalize across domains.

\section*{Acknowledgements}
We thank Xiang Fan, Khyathi Chandu, and Stella Li for their valuable feedback. This work was supported by the Singapore National Research Foundation and the National AI Group in the Singapore Ministry of Digital Development and Information under the AI Visiting Professorship Programme (award number AIVP-2024-001), by the AI2050 program at Schmidt Sciences, NSF DMS-2134012, and ONR N00014-24-1-2207.

\bibliography{colm2025_conference}
\bibliographystyle{colm2025_conference}

\appendix
\lstdefinestyle{mystyle}{escapechar=@}

\lstset{
  basicstyle=\ttfamily,  %
  literate={``}{{\textquotedblleft}}2 %
   {''}{{\textquotedblright}}2 %
   {'}{{\textquotesingle}}1   %
   {`}{{\textasciigrave}}1    %
}

\section{Limitations}
\label{sec:limitations}
One limitation of our work comes from our definition of ground truth factuality.
Our definition of the factuality of a claim is tied to whether it is entailed by a gold document.
However, there are two potential issues with this definition.
The first is that our gold documents (which are Wikipedia pages) might not contain all of the information that the model was pretrained with, so a model could generate claims that are correct, but are unsupported by the gold documents.
These unverifiable claims are difficult to deal with and future work might extend the range of gold documents we consider.
The second potential issue with defining factuality in terms of whether a claim is entailed by a gold document is that our entailment function is not perfect---as we observed in Table \ref{tab:qual_analysis_detail}, some claims which are true and supported by the documents may be incorrectly predicted to be \texttt{unsupported}.
(These tend to more vague or require additional reasoning.)
We do benchmark our entailment model on biography data with human-labeled claims from \citep{Min_Krishna_Lyu_Lewis_Yih_Koh_Iyyer_Zettlemoyer_Hajishirzi_2023}, finding that Minicheck has an accuracy within 1 percentage point of GPT4o (83.10\% vs 84.31\%), but there is still room for improvement.
Developing and using better fast entailment models is an important step for future work. %

In addition to limitations with our definition of factuality, another limitation comes from compounding errors in our pipeline.
There can be errors in any of the steps in our data curation pipeline---sentence-segmentation can split at incorrect punctuation; atomization can introduce new claims or miss some; decontextualization and merging claims can add new information that is not supported.
In practice, we do not observe too many problems from these stages of the pipeline, but any error in one has the potential to propagate down to others and affect our results. See Appendix \ref{appendix:pipeline-validation} for validation of the components of our pipeline.

\section{Additional Dataset Details}
\label{appendix:dataset-details}
\textbf{Bios}: This dataset provides a list of names along with that person's  
    Wikipedia page as a reference. 
    We extract $\sim$9,000 entities from Wikidata that are instances of  ``human'' (\texttt{Q5}), and have an English Wikipedia page.
    We query models using the same prompt as \citet{Min_Krishna_Lyu_Lewis_Yih_Koh_Iyyer_Zettlemoyer_Hajishirzi_2023}: \textit{``Tell me a bio of } \texttt{[NAME]}\textit{''}.
    When computing labels using $\fdoc$ or for training $\fprobe$, we use the entire documents, however as the Wikipedia pages for many of these entities are quite long, when constructing $\gold$, we use the opening section of the Wikipedia page as the gold document.
    
    \textbf{Plots}: This dataset provides contains $\sim$3,000 media titles (books, movies, plays, etc.) along with their plot summaries extracted from Wikipedia by \citet{riedl-2017-wikiplots}. We prompt the model with the prompt: \textit{``Write a summary of the plot of the book or film called } \texttt{[TITLE]}\textit{. Do not provide any additional information about the context or background. Just write a summary of the plot.''}.
    When atomizing this dataset, the individual sentences can be difficult to understand on their own, so we decontextualize sentences before atomizing them (i.e., we attempt to make the sentences make sense even when separated from the rest of the document)~\citep{choi-etal-2021-decontextualization, gunjal2024molecular}. (See Appendix~\S\ref{appendix:prompt-decontext} for the prompt used.)
    We use Mistral-Instruct-7B-v0.2 and a separate set of in-context examples to perform the decontextualization.
    
    \textbf{Medical Terms}~\citep{wiki_medical_term}: This dataset provides a set of $\sim$800 specific medical terms (\textit{e.g., Lymphogranuloma venereum}) along with the corresponding Wikipedia articles as reference. We format a simple prompt ``\textit{Write an expert-level paragraph with }\texttt{[TERM]}\textit{ as a topic. Include numerical details.}'' to generate a long-form explanation of the medical term.

\section{Validation of Pipeline Components}
\label{appendix:pipeline-validation}
\subsection{Choice of Atomization Function}
\label{appendix:atomization-choice}
\arredit{We used Mistral-7B-Instruct-v0.2 for atomization because we wanted an efficient method to decompose sentences into individual claims. Prior work such as FActScore \citep{Min_Krishna_Lyu_Lewis_Yih_Koh_Iyyer_Zettlemoyer_Hajishirzi_2023} uses API-based models, but using them our case quickly becomes slow and expensive.
At the onset of this work, Mistral-7B-Instruct-v0.2 was one of the best 7B-scale models available, so it was used.
Future work can investigate other models, such as the models fine-tuned by \citet{song-etal-2024-veriscore}.

We validated our use of Mistral-7B-Instruct-v0.2 using data released by the authors of FActScore.
This data includes the raw outputs from ChatGPT, InstructGPT, and Perplexity.ai as well as the atoms generated by InstructGPT.
We re-atomize these raw outputs using Mistral---for each sentence in the output, we extract its atomic claims, and compute the RougeL and exact-match scores between each Mitral atom and each InstructGPT atom for that sentence.
The score for that claim is the maximum of the RougeL and exact match scores, which accounts for any atoms that are combined or separated by either of the atomizers.
(We compute these matches for 157 of the 183 outputs where our method extracts the same number of sentences as theirs.)
The RougeL scores are quite high, indicating that Mistral and InstructGPT use similar $n$-grams in their atoms.
The exact match scores are lower, but comparable to the similarties reported in recent work \citep{song-etal-2024-veriscore}
(See Table \ref{appendix:tab-atomization-comp}).}

\begin{table}
    \centering
    \begin{tabular}{ccc}
        \toprule
         Generation Model & Max RougeL & Max EM \\
         \midrule
         ChatGPT & 0.838 & 0.429 \\
         InstructGPT & 0.852 & 0.486 \\
         Perplexity.ai & 0.802 & 0.399 \\
         \bottomrule
    \end{tabular}
    \caption{Maximum RougeL and exact match (EM) between atoms from Mistral-7B (ours) and InstructGPT (used by FActScore) for the outputs from three systems in the biography domain.}
    \label{appendix:tab-atomization-comp}
\end{table}

\subsection{Choice of Entailment Function}
\label{appendix:entailment-comp}
In our experiments, we used Minicheck (a fine-tuned flan-t5-large model) as our entailment model as it was accurate and fast \citep{tang2024minicheck}.
We validated this choice by comparing to other entailment functions in the biographies setting using approximately 3,000 of the human-generated and labeled-atomic claims collected by \citet{Min_Krishna_Lyu_Lewis_Yih_Koh_Iyyer_Zettlemoyer_Hajishirzi_2023} for 100 different entities.

We compared Minicheck to the following entailment models:
\begin{itemize}
    \item \textbf{Flan-T5-11B}: Treating fact verification as a mulitple choice task with two options: ``(A) Yes'' and ``(B) No. \citet{Guan_Dodge_Wadden_Huang_Peng_2023} found this model to perform well and we use their prompt.
    \item \textbf{Mistral}: Mistral-7B-Instruct-v0.2 is prompted to generate ``True" if the claim is true based on a document and ``False'' otherwise using the same prompt as \citet{Min_Krishna_Lyu_Lewis_Yih_Koh_Iyyer_Zettlemoyer_Hajishirzi_2023}.
    \item \textbf{Llama-Chat+NPM}: Llama-2-Chat and a nonparametric model are combined as in \citet{Min_Krishna_Lyu_Lewis_Yih_Koh_Iyyer_Zettlemoyer_Hajishirzi_2023}, using the same prompt as Mistral.
    \item \textbf{Llama-Chat+NPM Logits}: Similar to Llama-Chat+NPM, except the probability mass on the tokens corresponding to ``True'' and ``False'' are compared rather than just using the text that model generated.
    \item \textbf{GPT-4o, GPT-3.5-Turbo}: Use the same prompt as Mistral. ``True'' or ``False'' is determined by parsing the generation.
    \item \textbf{All True}: A baseline that predicts all claims are true.
\end{itemize}

The results are in Table \ref{appendix:tab-entailment-comp}. Minicheck performs only a little worse compared to GPT-4o but is much faster and cheaper.

\begin{table}
    \centering
    \begin{tabular}{ccc}
        \toprule
         Entailment Model & Macro F1 & Accuracy \\
         \midrule
         All True Baseline & 0.364 & 0.574 \\
         \midrule
         Llama-Chat+NPM Logits & 0.692 & 0.692 \\
         Llama-Chat+NPM & 0.731 & 0.732 \\
         Mistral & 0.740 & 0.754 \\
         Flan-T5-11B & 0.802 & 0.803 \\
         Minicheck & \textit{0.827} & \textit{0.831} \\
         \midrule
         GPT-3.5-Turbo & 0.731 & 0.754 \\
         GPT-4o & \textbf{0.840} & \textbf{0.843} \\
         \bottomrule
    \end{tabular}
    \caption{Comparison between entailment functions. \textbf{Minicheck performs comparably to GPT-4o but is much faster and cheaper.}}
    \label{appendix:tab-entailment-comp}
\end{table}

\subsection{Choice of Merging Function}
\label{appendix:merging-choice}

\arredit{We chose to use Llama-3-8B-Instruct for merging claims after manually comparing its outputs to those of Mistral-7B-Instruct-v0.2 and Llama-2-7B-chat.
In particular, we noticed that the latter two tended to omit claims when merging while Llama-3 did not.
There is also a risk of merging introducing hallucinations.
To investigate how often this occurs, we assess the factuality of the outputs of our synthetic data creation process.
If our atomization and merging systems are perfectly faithful, the factuality of the synthetic data should be 100\%.
We use our biography setting for this measurement.
Our training set of entities is quite large, so instead we run our evaluation entities through our synthetic data pipeline, and evaluate the outputs.
We compute this for Mistral and Llama-2 and find that the Mistral data has a factuality of 94.4\% and the Llama-2 data has a factuality of 92.7\%. 
We did a qualitative evaluation of 20 randomly sampled samples with errors to assess their causes. We found that 5 were due to merging introducing hallucinations. Our conclusion from this is that while merging occasionally introduces hallucinations, the are relatively rare.}

\section{Prompts}
\subsection{Atomization}
\label{appendix:prompt-atomize}
Below is the template for the prompt used for atomizing generated sentences. We used Mistral-7B-Instruct-v0.2 with temperature=0.2 to perform atomization.
\begin{lstlisting}[breaklines=true, style=mystyle]
[Examples x8]
Please breakdown the following sentence into independent facts: [Sentence]
- [claim 1]
- [claim 2]
- [...]

Please breakdown the following sentence into independent facts: [sentence]
\end{lstlisting}

\subsection{Dataset Construction}
\label{appendix:prompt-merge}
The prompt for merging atomic claims back into a coherent data sample is below. We used Llama-3-8B-Instruct with temperature=0.2 to perform merging:

\begin{lstlisting}[breaklines=true, style=mystyle]
Given a list of sentences, merge them into a natural paragraph. Do not add or remove any information, only write a paragraph based on the information from the list of sentences.

[Examples x3]
Sentences:
- [sentence 1]
- [sentence 2]
- ...
Merged paragraph: [merged paragraph]

...

Sentences:
- [claim 1]
- [claim 2]
- ...
Merged paragraph:
\end{lstlisting}

\subsection{Decontextualization Prompt for Plots Dataset}
\label{appendix:prompt-decontext}
The prompt for decontextualizing sentences in the plots dataset is below. In context examples includes resolving pronouns and other anaphora as well as examples that require no resolution.
We used Mistral-7B-Instruct-v0.2 with temperature=0.2 to perform decontextualization:
\begin{lstlisting}[breaklines=true, style=mystyle]
Rewrite each passage using its context.
[Examples x5]
Context: [Preceding Sentences]
Passage: [Sentence to decontextualize]
Rewrite: [Decontextualized sentences]

Context: [Preceding Sentences]
Passage: [Sentence to decontextualize]
Rewrite:
\end{lstlisting}

\section{Probe Details}
\label{appendix:probe-f1}
We train probes on claims from gold documents and generations on held out entities from our datasets.
The prompt we use to encode the claims is \texttt{[Prompt containing entity]: [Claim]}.
We found that including the prompt provided necessary context for the model and helped achieve sufficient probe performance.
Each probe is a linear classifier trained on a binary classification task, trained with no bias term and a maximum of 1,000 gradient steps.

\begin{figure}[ht]
    \centering
    \includegraphics[width=\linewidth]{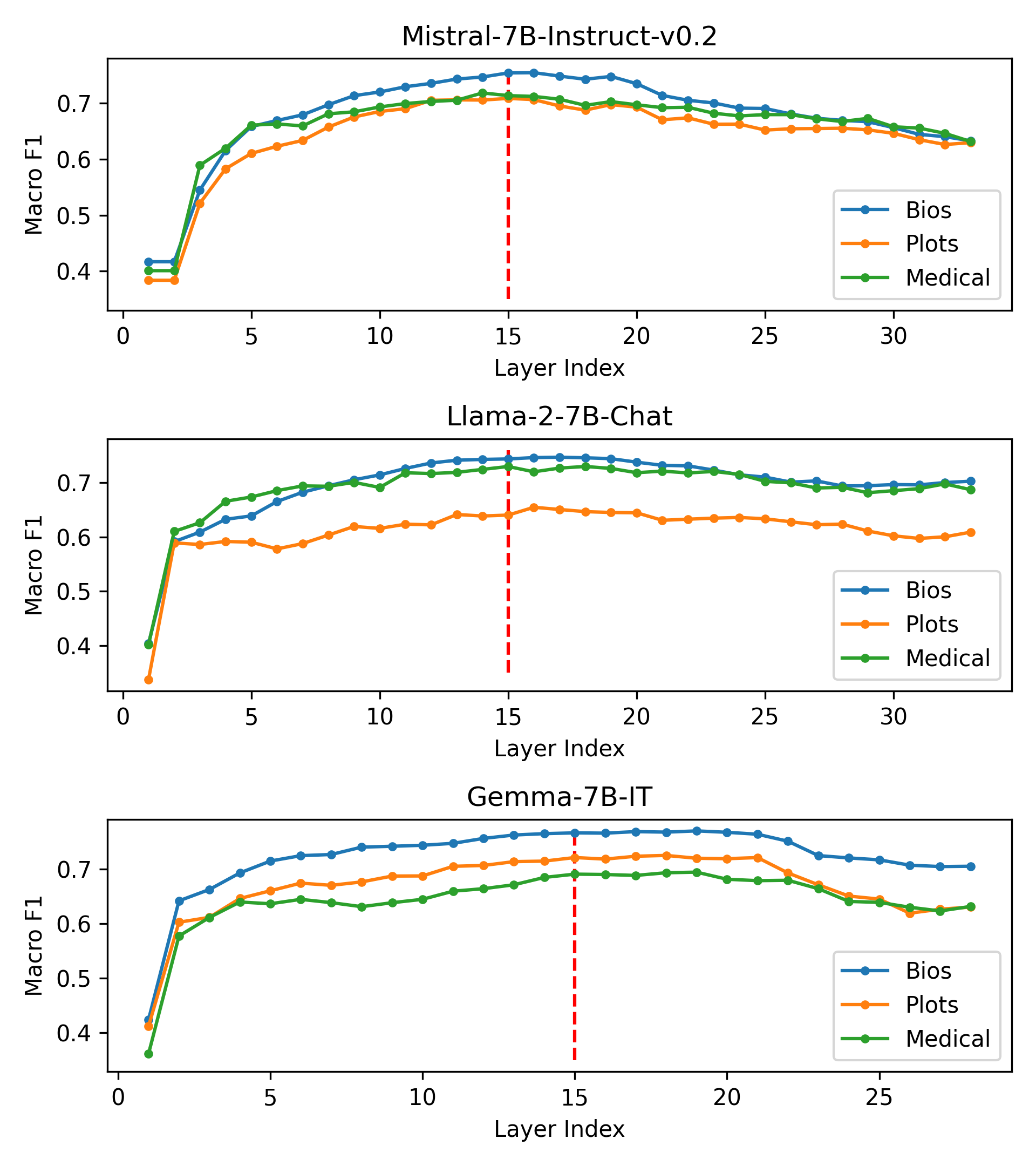}
    \caption{Plot of probe F1 versus layer of the model on claims from the evaluation set of model generations. The F1 score peaks in the early-middle layers of the model.
    The index we select for our experiments (15) is marked with the red dashed line. }
    \label{fig:appendix-probe-f1}
\end{figure}

When we plot probe performance against layer index, we find that probes at the middle layers of the last token index have the highest F1 scores (Figure \ref{fig:appendix-probe-f1}).

\section{Analysis of Training Data}
\subsection{Training Data Factuality}
\label{appendix:train-data-factuality}
We assess the factuality of our training data by computing the percentage of atoms entailed by gold documents remaining after our filtering steps.
By definition, training data filtered with external documents is considered to be 100\% factual. 
Training data filtered with $\fprobe$ includes fewer factual claims according to our data sources than that filtered with $\fdoc$, however leads to \textit{more} factual generations when used for training according to those same data sources (Table~\ref{appendix:tab-train-data-factuality}).

\subsection{Factuality vs Claim Position}
\label{appendix:factuality-vs-length}
\begin{figure*}
    \centering
    \includegraphics[width=\linewidth]{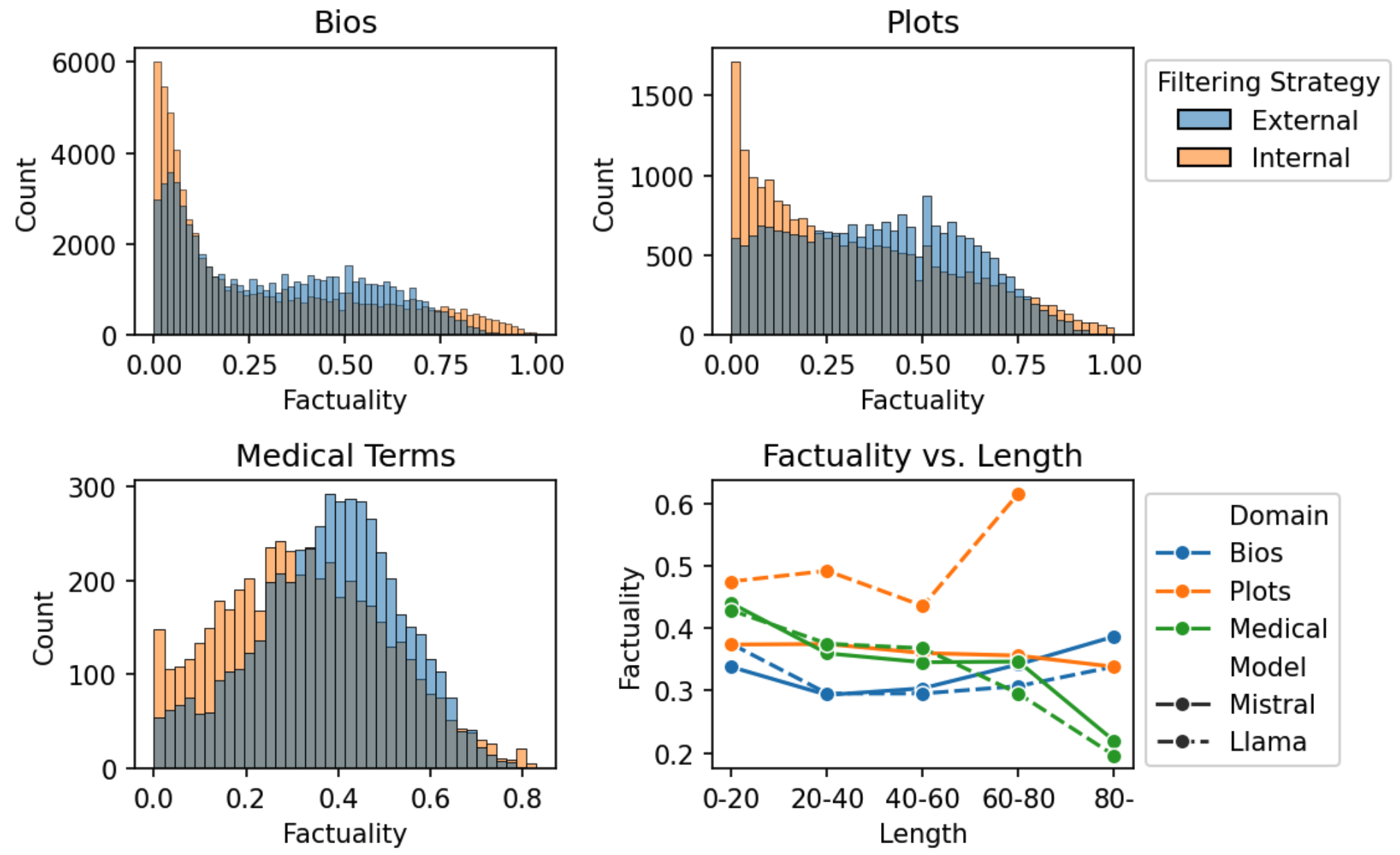}
    \caption{Histograms showing the proportion of claims predicted to be supported for each generation by each of the filtering strategies (for Mistral). Bottom right is a plot of factuality (measured with entailment) versus the position of the fact in the document. (This plot was created using the training data generated by each of the models for each of the datasets.) \textbf{For the three domains we study, $\fprobe$ predicts fewer claims are supported than $\fdoc$.}}
    \label{appendix:fig-factuality-int-vs-ext}
\end{figure*}

We find that the factuality of model generations vary with their position in the generations (different from prior work, we find that whether higher positions are more or less factual somewhat depends on the domain).
This is potentially problematic---if the different filtering methods remove different proportions of claims, the length of the resulting generations can confound our results.
We do in fact find that $\fprobe$ tends to predict that more atomic claims are unsupported compared to $\fdoc$ (Figure~\ref{appendix:fig-factuality-int-vs-ext}).
This motivates controlling for length in our experiments.

\begin{table}[t]
\small
\renewcommand*{\arraystretch}{1.2}
\centering
\begin{tabular}{cc|ccc}
    \toprule
    \multicolumn{2}{c|}{\textbf{Data Construction Method}} 
    & \multicolumn{3}{c}{\textbf{Factuality} ($\uparrow$)} \\
    Domain & Filtering Strategy & Mistral & Llama 2 & Gemma 2\\
    \midrule
    \multirow[c]{2}{*}{\textit{Plots}} & \textit{External} & 100 & 100 & 100\\
     & \textit{Internal} & 84.2 & 72.8 & 63.5\\
     \midrule
     \multirow[c]{2}{*}{\textit{Bios}} & \textit{External} & 100 & 100 & 100 \\
     & \textit{Internal} & 85.6 & 79.2 & 85.1\\
     \midrule
     \multirow[c]{2}{*}{\textit{Medical}} & \textit{External} & 100 & 100 & 100\\
     & \textit{Internal} & 79.8 & 75.5 & 76.9 \\
     \bottomrule
\end{tabular}
\caption{The percentage of atoms that are factual after filtering. The model-generated training data filtered internal probes includes claims that are not supported by gold documents, whereas all of the claims in the training data filtered with external document knowledge are supported.}
\label{appendix:tab-train-data-factuality}
\end{table}

\subsection{Internal vs External Filtering}
\label{appendix:internal-vs-external-filtering}
\begin{table}[t]
    \renewcommand{\arraystretch}{1.4}
    \small
    \centering
    \begin{tabular}{@{\hskip .1em}p{2em}p{20em}}
         \toprule
         \multicolumn{2}{c}{\underline{George Washington}}\\
         \multirow[c]{3}{*}{$\fprobe$} & He was taught the classics. \\
         & His mother remarried. \\
         & He later dropped his maternal surname. \\
         \cmidrule{2-2}
         \multirow[c]{3}{*}{$\fdoc$} & Washington was born at Popes Creek Landing.\\
         & He lived on his family's plantation once more after 1752. \\
         &  The Department of State was established during his presidency. \\
         \midrule
         \multicolumn{2}{c}{\underline{Karl Georg von Raumer}} \\
         \multirow[c]{3}{*}{$\fprobe$} & Von Raumer died. \\
         & Raumer published books. \\
         & Raumer was born in Germany. \\
         \cmidrule{2-2}
         \multirow[c]{3}{*}{$\fdoc$} & Karl Georg von Raumer was a geologist. \\
         & His major works include "Lehrbuch der Geologie". \\
         & He was made a professor at the University of Halle. \\
         \bottomrule
    \end{tabular}
    \caption{Selected examples of claims that are predicted to be true by only $\fprobe$ and only $\fdoc$ for common (Washington) and rarer (von Raumer) entities. \textbf{Overall, $\fprobe$'s predictions are less detailed than $\fdoc$'s}.
    }
    \label{tab:qual_analysis_detail}
\end{table}

We find that when $\fprobe$ and $\fdoc$ disagree on a claim, the claims tends to be shorter when predicted to be supported by $\fprobe$ and longer when predicted to be supported by $\fdoc$.
We compute the number of white-space separated tokens in claims taken from the model generations predicted to be supported by $\fprobe$ and $\fdoc$, and find that $\fprobe$'s are shorter in all cases (Table~\ref{appendix:tab-internal-vs-external-filtering}).
Qualitative examples of the difference between the facts assigned different labels by each filter are in Table \ref{tab:qual_analysis_detail}.

\begin{table}[t]
\renewcommand*{\arraystretch}{1.2}
\small
\centering
\begin{tabular}{cc|cc}
\toprule
Model & Domain &  $\fprobe$ & $\fdoc$   \\
\midrule
\multirow[c]{3}{*}{\textit{Gemma}} & \textit{Bios} & 5.68 & 6.26  \\
 & \textit{Plots} & 6.67 & 7.11 \\
 & \textit{Medical} & 6.81 & 6.99 \\
\multirow[c]{3}{*}{\textit{Llama-2}} & \textit{Bios} & 5.83 & 7.46 \\
 & \textit{Plots} & 7.48 & 7.95 \\
 & \textit{Medical} & 7.63 & 7.75 \\
\multirow[c]{3}{*}{\textit{Mistral}} & \textit{Bios} & 5.83 & 7.02 \\
 & \textit{Plots} & 5.77 & 7.03 \\
 & \textit{Medical} & 7.02 & 7.93 \\
\bottomrule
\end{tabular}
\caption{Lengths of claims assigned the label ``supported'' by $\fprobe$ and $\fdoc$ respectively. In general, $\fprobe$'s claims tend to be shorter. In all cases, $\fprobe$'s claims are significantly shorter than $\fdoc$'s (with significance determined by a t-test for related samples and $p < 0.05$.)}
\label{appendix:tab-internal-vs-external-filtering}
\end{table}

\subsection{Diversity}
\label{appendix:diversity}
We measured diversity of the training data using common metrics such as compression ratio (CR) (the ratio of the uncompressed text to text compressed with gzip; higher is less diverse) and n-gram diversity (the fraction of unique $n$-grams \cite{meister-etal-2023-locally}---we report $n=4$ (4GD); higher is more diverse) using the \texttt{diversity} package \citep{shaib2024Diversity}.
See results in Table \ref{appendix:tab-diversity}.
In general, the generated training data is less diverse, and filtering with $\fprobe$ leads to less diverse training data.
\begin{table*}[t]\centering
\resizebox{.95\textwidth}{!}{
    \begin{tabular}{ccc|cccc}\toprule
        \multicolumn{3}{c|}{\textbf{Data Construction Method}} & \multicolumn{2}{c}{\textbf{Mistral-7B-Instruct}} & \multicolumn{2}{c}{\textbf{Llama-2-7B-Chat}} \\ \cmidrule(lr){1-3} \cmidrule(lr){4-5} \cmidrule(lr){6-7}
         Domain & Knowledge Source & Filtering Strategy & CR ($\downarrow$) & 4GD ($\uparrow$) & CR ($\downarrow$) & 4GD ($\uparrow$) \\
         \midrule
         \multirow[c]{4}{*}{\textit{Plots}}
            & \multirow[c]{2}{*}{\textit{Gold}} & \textit{External} & 2.615 & 2.691 & 2.545 & 2.774 \\
            &  & \textit{Internal} & 2.846 & 2.584 & 2.626 & 2.724 \\ \cmidrule(lr){2-7}
            & \multirow[c]{2}{*}{\textit{Generated}} & \textit{External} & 2.875 & 2.537 & 3.084 & 2.328\\
            &  & \textit{Internal} & 2.981 & 2.477 & 3.128 & 2.314 \\
        \midrule
        \multirow[c]{4}{*}{\textit{Bios}}
            & \multirow[c]{2}{*}{\textit{Gold}} & \textit{External} & 2.786 & 2.266 & 2.661 & 2.295 \\
            &  & \textit{Internal} & 2.848 & 2.269 & - & - \\ \cmidrule(lr){2-7}
            & \multirow[c]{2}{*}{\textit{Generated}} & \textit{External} & 2.966 & 2.122 & 3.194 & 1.824 \\
            &  & \textit{Internal} & 3.107 & 2.035 & 3.325 & 1.75 \\
        \midrule
        \multirow[c]{4}{*}{\textit{Medical}}
            & \multirow[c]{2}{*}{\textit{Gold}} & \textit{External} & 3.049 & 2.664 & 2.927 & 2.706 \\
            &  & \textit{Internal} & 3.178 & 2.576 & 2.92 & 2.714\\ \cmidrule(lr){2-7}
            & \multirow[c]{2}{*}{\textit{Generated}} & \textit{External} & 3.243 & 2.577 & 3.466 & 2.397 \\
            &  & \textit{Internal} & 3.371 & 2.505 & 3.522 & 2.359 \\
        \bottomrule
    \end{tabular}
}
\caption{Diversity results from the training data for various settings Mistral-7B-Instruct-v0.2 and Llama 2 Chat. CR = compression raito and NGD is n-gram diversity (where $n=4$).}
\label{appendix:tab-diversity}
\end{table*}

\section{Additional Training Details}
\label{appendix:hyperparam-search}
For our hyperparameter search, we varied the number of training steps (ranging from 500 to 1000), the peak learning rate (1e-4, 3e-4, 5e-4), and the lora rank (8, and 128).
We selected the 500 training steps, a peak learning rate of 3e-4, and a lora rank of 8 based on performance on a held out set of biography data.
We train using NVIDIA A100 and A40 gpus.

\section{Additional Results and Analysis}
\subsection{Factuality vs Frequency}
\label{appendix:frequency}
\begin{figure}
    \centering
    \includegraphics[width=0.6\linewidth]{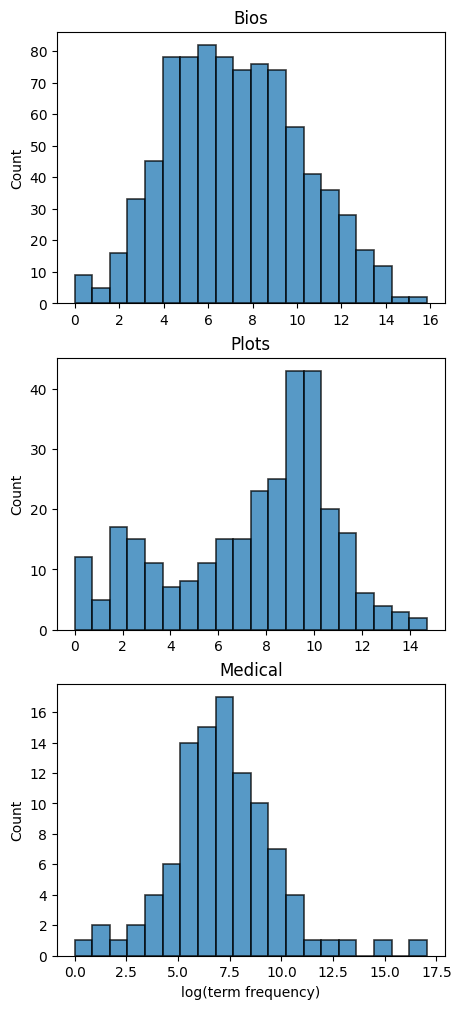}
    \caption{Histogram of the log-frequencies of the terms in our evaluation datasets.}
    \label{appendix:fig-freq-hist}
\end{figure}

Prior work has argued that a models' familiarity with a term is correlated with its frequency \citep{Min_Krishna_Lyu_Lewis_Yih_Koh_Iyyer_Zettlemoyer_Hajishirzi_2023, Mallen_Asai_Zhong_Das_Khashabi_Hajishirzi_2023a}.
Due to this, we would expect that the inputs whose factuality are most improved by \analysistool{} are ones that are rarer.
To determine the extent to which this is the case, we use the infinigram api \cite{Liu2024InfiniGram} to obtain the number of occurrences of each of our evaluation terms in the red-pajama dataset (a representative pre-training dataset) \citep{weber2024redpajama}.
Histograms of the log-frequencies of these terms are in Figure \ref{appendix:fig-freq-hist}.

We then compute the difference between the factuality of models trained with model-generated data filtered using $\fprobe$ and $\fdoc$ on the same terms ($\fprobe - \fdoc$).
We expect these differences to be higher for the rare terms and lower for the more frequent terms.
We bucket these terms in quartiles by frequencies and compute the mean of the factualities in each quartile, visible in Table \ref{table:factuality-vs-frequency}.
We observe in general that the greatest benefits come in the bottom quartiles, except in the medical domain, where the benefits are observed to be more uniform.

\subsection{TriviaQA results}
In our work, we focus on long-form generation tasks.
However, the settings we test can also be extend to short-form generation tasks.
We test whether factual training data is necessary for improving performance on the short-form QA task, the wikipedia split of TriviaQA \cite{JoshiTriviaQA2017}, using Mistral-7B-Instruct and Llama-2-7B-Chat (Table \ref{tab:triviaqa_results_table}). We find that internal filtering narrowly under-performs external filtering in this setting, indicating that factuality might matter more in this setting.

\begin{table*}[t]\centering
\resizebox{.95\textwidth}{!}{
    \begin{tabular}{ccc|cccccc}\toprule
        \multicolumn{3}{c|}{\textbf{Data Construction Method}} & \multicolumn{3}{c}{\textbf{Mistral-7B-Instruct}} & \multicolumn{3}{c}{\textbf{Llama-2-7B-Chat}} \\ \cmidrule(lr){1-3} \cmidrule(lr){4-6} \cmidrule(lr){7-9}
         Domain & Knowledge Source & Filtering Strategy & Factuality ($\uparrow$)& Detail ($\uparrow$) & Abstentions ($\downarrow$) & Factuality ($\uparrow$) & Detail ($\uparrow$) & Abstentions ($\downarrow$) \\
         \midrule
         \multirow[c]{6}{*}{\textit{TriviaQA}} & \textit{None} & \textit{None} & 0.683 & 1.000 & 0.000 & 0.602 & 1.000 & 0.000 \\ \cmidrule(lr){2-9}
            & \multirow[c]{2}{*}{\textit{Gold}} & \textit{External} & 0.442$_{\text{(0.009)}}$ & 1.000$_{\text{(0.000)}}$ & 0.000$_{\text{(0.000)}}$ & 0.398$_{\text{(0.004)}}$ & 1.000$_{\text{(0.000)}}$ & 0.000$_{\text{(0.017)}}$ \\ 
            & & \textit{Internal} & 0.562$_{\text{(0.004)}}$ & 1.000$_{\text{(0.000)}}$ & 0.425$_{\text{(0.021)}}$ & 0.535$_{\text{(0.014)}}$ & 1.000$_{\text{(0.000)}}$ & 0.739$_{\text{(0.000)}}$ \\ \cmidrule(lr){2-9}
            & \multirow[c]{2}{*}{\textit{Generated}} & \textit{External} & \textbf{0.729}$_{\text{(0.047)}}$ & 1.000$_{\text{(0.000)}}$ & 0.190$_{\text{(0.042)}}$ & \textbf{0.658}$_{\text{(0.008)}}$ & 1.000$_{\text{(0.000)}}$ & 0.540$_{\text{(0.064)}}$ \\
            & & \textit{Internal} & \cellcolor{\ccolor} 0.710$_{\text{(0.064)}}$ & \cellcolor{\ccolor} 1.000$_{\text{(0.000)}}$ & \cellcolor{\ccolor} 0.218$_{\text{(0.019)}}$ & \cellcolor{\ccolor} 0.648$_{\text{(0.008)}}$ & \cellcolor{\ccolor} 1.000$_{\text{(0.000)}}$ & \cellcolor{\ccolor} 0.607$_{\text{(0.023)}}$ \\
        \bottomrule
    \end{tabular}
}

\caption{Main results from fine-tuning Mistral-7B-Instruct-v0.2 and Llama-2-7B-Chat on TriviaQA (reading comprehension, Wikipedia, no context split). Numbers in parentheses are standard deviations across three random seeds for all sampling done in the pipeline. Dashes are where the Bios internal knowledge filter for Llama-2-7B-Chat predicts all claims are \texttt{unsupported}. The random filtering baseline randomly keeps the same number atoms as the other filtering strategies without considering their factuality \textbf{Using Generated with $\fprobe$ consistently leads to the highest factuality.} Factuality is percentage of correct claims, detail is average number of claims in non-abstaining responses, and abstentions is the percentage of abstaining responses.
Shaded rows represent \analysistool{}.
}

\label{tab:triviaqa_results_table}
\end{table*}

\subsection{Human Analysis of Generation Detail}
\label{appendix:detail_analysis}
In our work, we focus on the number of claims in a generation as a measure of its detail; however, this metric says nothing about the informativity of these claims.
Here, we conduct a qualitative analysis to estimate whether the generations from models trained with \analysistool{} are still informative.
We conduct this analysis using generations from the Mistral model finetuned with \analysistool{} in the biographies dataset setting.
We randomly sample 100 entities and select the first generation from each.
We have two annotators manually rate the level of detail of a generation on a three point scale without considering factuality:
\begin{itemize}
    \item A \textbf{score of 0} is assigned to generations that provide no additional non-trivial information or abstain.
    \item A \textbf{score of 1} is assigned to generations that provide vague or disfluent responses. There may be some details, but there are still gaps in the generation that the reader wants filled.
    \item A \textbf{score of 2} is assigned to generations that provide a solid response containing many details.
\end{itemize}
See Table \ref{appendix:tab-detail} for examples.

The inter-annotator agreement was measured using Cohen's $\kappa$, and was $0.79$.
Overall, the average score assigned to the generations was $0.95$, and if the 17 abstaining responses are omitted, the average score was $1.145$, with both annotators giving a score of 1 or higher to 71\% of the generations (and a score of 2 to 37\%).

We take these results to indicate that the generations from models trained with \analysistool{} do contain some details, but are more vague overall compared to the baseline generations (which are always quite detailed).
Vague generations are not necessarily bad---they represent one strategy to increase the factuality of models separate from abstaining outright.
In particular, they provide \textit{some} additional information about the entity, even if the information is incomplete.

\begin{table*}
    \renewcommand{\arraystretch}{1.4}
    \small
    \centering
    \begin{tabular}{p{40em}}
    \toprule
        \textbf{Score of 0}\\
        Kurt Eberhard came to prominence. \\
        Gregor Wentzel was born and is known for his work. \\
        \midrule
        \textbf{Score of 1}\\
         Tarateño Rojas, a renowned Argentinian performer, is a musician and showman who has made his career and is widely regarded. \\
         Max Schreck began his career in the theater, making a transition to film. \\
         \midrule
         \textbf{Score of 2} \\
         Nicolas Poussin, a French painter, was one of the most influential artists of the seventeenth century. Born in Paris, France, Poussin's early work was admired, and he spent several years studying and traveling throughout France and Italy. In Rome, Italy, where he spent the rest of his life, Poussin began to develop his distinctive style, characterized by a classical, orderly approach to composition and subject matter. His paintings, which often depicted scenes from ancient mythology, the Bible, or classical history, were carefully planned and executed, and he is known for his mastery of figure painting, color, and the use of light and shadow. Poussin's work was significant during his lifetime and remains highly regarded today, influencing many artists and continuing to inspire artists in the centuries since his death. \\
         Emily Warren Roebling, an American, displayed remarkable intelligence and persevered during a period when women were discouraged from pursuing careers in engineering. She studied engineering and her efforts earned her the moniker "Mother of the Brooklyn Bridge" because of her vital role during the construction of the Brooklyn Bridge. Roebling, wife of Washington Roebling, passed away at the age of 77, leaving behind a lasting legacy as a pioneering woman in engineering. \\
         \bottomrule
    \end{tabular}
    \caption{Examples of generations that fall into our three score buckets for assessing the detail of model generations. A score of 0 is assigned to generations that provide no additional non-trivial information or abstain. A score of 1 is assigned to generations that provide vague or disfluent responses. There may be some details, but there are still gaps in the generation that the reader wants filled. A score of 2 is assigned to generations that provide a solid response containing many details. Generations are generated by Mistral-7B-Instruct-v0.2 finetuned with \analysistool{}.}
    \label{appendix:tab-detail}
\end{table*}

\section{Sample Data}
\label{appendix:sample-data}
For examples of the data samples, see Table \ref{appendix:tab-sample-data} for samples filtered with internal knowledge, Table \ref{appendix:tab-sample-data-ext} for samples filtered with external knowledge, and Table \ref{appendix:tab-sample-data-medical} for samples from the Medical domain. Overall, the styles of the two knowledge sources are similar.
Finally, for an example of generations before and after fine-tuning, see Table \ref{appendix:tab-sample-generations}. All data samples are filtered and generated using Mistral-7B-Instruct-v0.2.

\begin{table*}
    \small
    \centering
    \begin{tabular}{@{}p{350pt}@{}} 
    \toprule
    Topic: \textbf{Barack Obama}, Knowledge Source: Gold, Filtering Strategy: Internal 
         \\
    \rule{0pt}{2ex}
         Barack Obama, an American and former politician, served as the 44th president of the United States from 2009 to 2017. Born on August 4, 1961, in Honolulu, Hawaii, Obama was the first African-American president of the United States. Prior to his presidency, he served as a U.S. senator representing Illinois and worked as a community organizer in Chicago. Obama received national attention in 2004 after his speech at the Democratic National Convention, and he was nominated by the Democratic Party for president in 2008. He chose Joe Biden as his running mate and was elected over John McCain. During his first term, Obama passed major legislation, including a stimulus package, health care legislation, and financial regulation reform. He also appointed two Supreme Court justices, Sonia Sotomayor and Elena Kagan. Obama won re-election in 2012, defeating Republican opponent Mitt Romney, and served a second term as President. During his second term, Obama presided over the implementation of the Affordable Care Act and other legislation passed in his first term.\\

    \midrule
         Topic: \textbf{Barack Obama}, Knowledge Source: Generated, Filtering Strategy: Internal \\ 
    \rule{0pt}{2ex}
Barack Obama, the 44th president of the United States and the first African American to hold the office, was born in Honolulu, Hawaii. He attended Columbia University and earned his law degree from Harvard Law School. Obama worked as a community organizer and civil rights lawyer in Chicago before delivering the keynote address at the 2004 Democratic National Convention, which catapulted him into the national spotlight. He went on to serve in the U.S. Senate from 2005 to 2008 and announced his candidacy for the presidency in 2008, winning the Democratic nomination and going on to become the President. During his time in office, Obama passed several landmark policies, including the Affordable Care Act (Obamacare) and the Dodd-Frank Wall Street Reform and Consumer Protection Act, and oversaw the withdrawal of U.S. troops from Iraq. After leaving office, Obama and his family moved to Washington D.C., where they have remained active in politics and public service. \\
\midrule
         
    Topic: \textbf{Animal Farm}, Knowledge Source: Gold, Filtering Strategy: Internal \\
    \rule{0pt}{2ex}
    The animals, led by Snowball and Napoleon, revolt against Mr. Jones and drive him from the farm. The pigs, Snowball and Napoleon, elevate themselves to positions of leadership, and Napoleon eventually declares himself leader, implementing changes to the governance structure of Animal Farm. Napoleon appears to be adopting the lifestyle of a man, and the animals remain convinced of his leadership. However, despite the initial ideals, Napoleon advocates for the pigs' own interests, and the pigs begin to resemble humans. The animals observe the pigs, who have abolished revolutionary traditions and are now indistinguishable from the humans they initially rebelled against.\\
    \midrule
    Topic: \textbf{Animal Farm}, Knowledge Source: Generated, Filtering Strategy: Internal\\
    \rule{0pt}{2ex}
    "Animal Farm" is an allegorical tale about a group of farm animals who rebel against their human farmer, aiming to create a society where they can be equal, free, and in charge. The pigs lead the revolution, renaming the farm "Animal Farm" and establishing the Seven Commandments for the newly formed community. Initially, the rebellion succeeds, and the animals have control over their living conditions. However, the pigs gradually gain more power, twisting the rules and instituting human-like hierarchies and privileges. The other animals trust their leaders, unaware that the pigs are manipulating their power and growing increasingly despotic. As the situation worsens, Snowball is expelled due to fabricated allegations, and the animals come to a shocking realization: their revolution has resulted in a new tyrant, a far cry from the fair and just society they had envisioned.\\
    \bottomrule
    \end{tabular}
    \caption{Samples from training data filtered with internal knowledge probes from each of the Bios, Plots, and Medical domains respectively. We do not observe notable quality differences between the knowledge sources.}
    \label{appendix:tab-sample-data}
\end{table*}

\begin{table*}
    \small
    \centering
    \begin{tabular}{@{}p{400pt}@{}}
    \toprule
        Topic: \textbf{Barack Obama}, Knowledge Source: Gold, Filtering Strategy: External  \\
        \rule{0pt}{2ex}
  Barack Obama, an American and former politician, served as the 44th president of the United States from 2009 to 2017. Born on August 4, 1961, in Honolulu, Hawaii, Obama graduated from Columbia University in 1983 and later earned a law degree from Harvard Law School. He worked as a community organizer in Chicago and later taught constitutional law at the University of Chicago Law School. Obama entered elective politics in 1997, representing the 13th district in the Illinois Senate until 2004. He won the March Senate primary and delivered a well-received keynote address at the Democratic National Convention in July. Obama was elected to the Senate in November 2004 and was nominated by the Democratic Party for president in 2008. He chose Joe Biden as his running mate and was elected over John McCain, becoming the first African-American president of the United States. \\
    \midrule
    Topic: \textbf{Barack Obama}, Knowledge Source: Generated, Filtering Strategy: External \\
    \rule{0pt}{2ex}
    Barack Obama, the 44th president of the United States and the first African American to hold the office, was born in Honolulu, Hawaii. He attended Columbia University and earned his law degree from Harvard Law School. Obama worked as a community organizer and civil rights lawyer in Chicago before delivering the keynote address at the 2004 Democratic National Convention, which catapulted him into the national spotlight. He went on to serve in the U.S. Senate representing Illinois from 2005 to 2008 and then announced his candidacy for the presidency, winning the Democratic nomination in 2008. Obama went on to become the President, defeating Republican nominee John McCain, and served two terms from 2009 to 2017. During his time in office, he passed several landmark policies, including the Affordable Care Act (Obamacare) and the Dodd-Frank Wall Street Reform and Consumer Protection Act. Obama oversaw the withdrawal of U.S. troops from Iraq and increased troop levels in Afghanistan, while also pursuing diplomacy with countries such as Iran and Cuba. After leaving office, Obama and his family moved to Washington D.C., where they have remained active in politics. He has written several books, launched the Obama Foundation, which focuses on increasing opportunities for young people, and continues to be a prominent figure in American politics, advocating for progressive values and holding significant influence on a global scale.\\
    \midrule
    Topic: \textbf{Animal Farm}, Knowledge Source: Gold, Filtering Strategy: External  \\
    \rule{0pt}{2ex}
    Old Major, a wise old boar on the Manor Farm, summons the animals together for a meeting and teaches them a revolutionary song called \"Beasts of England\". After his death, Snowball and Napoleon assume command and consider it their duty to prepare for the Rebellion, which they were instructed by Old Major. The animals revolt, driving Mr. Jones, a drunken and irresponsible human, from the farm and renaming it "Animal Farm". Snowball and Napoleon lead the animals, adopting the Seven Commandments of Animalism, with the most important being "All animals are equal". Snowball teaches animals to read and write on the principles of Animalism, while Napoleon educates young puppies on the principles of Animalism.\\
    \midrule
    Topic: \textbf{Animal Farm}, Knowledge Source: Generated, Filtering Strategy: External  \\
    \rule{0pt}{2ex}
    "Animal Farm" is an allegorical tale about a group of farm animals who rebel against their human farmer, aiming to create a society where they can be equal, free, and in charge. The pigs lead the revolution, renaming the farm "Animal Farm" and establishing the Seven Commandments for the newly formed community. Initially, the rebellion succeeds, and the animals have control over their living conditions. However, the pigs gradually gain more power, twisting the rules and instituting human-like hierarchies and privileges. The other animals trust their leaders to uphold a fair and just society, but the pigs manipulate their power, growing ever more despotic. As the situation at Animal Farm worsens, the animals come to a shocking revelation: the pigs have become indistinguishable from the former human rulers, and their revolution has failed to achieve its original goals.\\
  \bottomrule
    \end{tabular}
    \caption{Samples from training data filtered with external fact verifiers from each of the Bios and Plots domains respectively. We do not observe notable quality differences between the knowledge sources.}
    \label{appendix:tab-sample-data-ext}
\end{table*}

\begin{table}
    \small
    \centering
    \begin{tabular}{@{}p{350pt}@{}} 
    \toprule
    Topic: \textbf{Actinic keratosis}, Knowledge Source: Gold, Filtering Strategy: Internal\\
    \rule{0pt}{2ex}
    Actinic keratosis, also known as solar keratosis or senile keratosis, is a pre-cancerous disorder characterized by thick, scaly, or crusty areas of skin. It is induced by ultraviolet light exposure, which causes damage to the skin, leading to the formation of these growths. These growths are more common in fair-skinned people and those who are frequently in the sun. If left untreated, actinic keratosis lesions have a risk of progression to squamous cell carcinoma, a type of skin cancer, making treatment by a dermatologist recommended.\\
    \midrule
    Topic: \textbf{Actinic keratosis}, Knowledge Source: Generated, Filtering Strategy: Internal\\
    \rule{0pt}{2ex}
    Actinic keratoses (AK) are typically the result of chronic sun exposure and are considered precancerous, having the potential to progress into squamous cell carcinoma. The prevalence of AK increases with age and typically occurs on sun-exposed areas such as the face. The diagnosis of AK is made through a physical examination of the skin, and treatment options are available. To prevent AK, regular sun protection measures are crucial, including wearing protective clothing, hats, and using sunscreen.\\
     \midrule
    Topic: \textbf{Actinic keratosis}, Knowledge Source: Gold, Filtering Strategy: External  \\
    \rule{0pt}{2ex}
    Actinic keratosis, also known as solar keratosis or senile keratosis, is a pre-cancerous disorder characterized by thick, scaly, or crusty areas of skin. It is induced by ultraviolet light exposure, which damages the skin and causes these growths to form over the course of decades. These growths are more common in fair-skinned people and those who are frequently in the sun, and are believed to form when skin gets damaged by UV radiation. If left untreated, actinic keratosis lesions have a risk of up to 20\% of turning into squamous cell carcinoma, a type of skin cancer.\\
    \midrule
    Topic: \textbf{Actinic keratosis}, Knowledge Source: Generated, Filtering Strategy: External  \\
    \rule{0pt}{2ex}
    Actinic Keratosis (AK) lesions are typically the result of chronic sun exposure and are considered precancerous, having the potential to progress into squamous cell carcinoma. These lesions occur on sun-exposed areas and are diagnosed through a physical examination of the skin by a healthcare professional. Treatment options are available for AK, including topical therapies such as 5-fluorouracil and imiquimod, as well as cryotherapy and surgical excision. Regular sun protection measures, such as using sunscreen, are crucial to prevent the development of AK lesions.\\
    \bottomrule
    \end{tabular}
    \caption{Samples from training data filtered with external fact verifiers and internal probes from the Medical domain. We do not observe notable quality differences between the knowledge sources.}
    \label{appendix:tab-sample-data-medical}
\end{table}

\begin{table}
    \small
    \centering
    \begin{tabular}{@{}p{350pt}@{}} 
    \toprule
    Topic: \textbf{Nicolas Sarkozy}, No Finetuning\\
    \rule{0pt}{2ex}
    Nicolas Sarkozy was born on January 28, 1955, in Paris, France. He served as the President of France from 2007 to 2012. Sarkozy's political career began in the center-right RPR party, where he rose quickly through the ranks, becoming its secretary-general in 1997. In 2002, he joined the Union for a Popular Movement (UMP), a new center-right party formed by then-Prime Minister Jean-Pierre Raffarin.\\
\\
Sarkozy first came to national prominence as the budget minister in Jacques Chirac's government from 2002 to 2004. He was known for his tough stance on crime and his efforts to reform the pension system. In 2004, Sarkozy was elected as the mayor of Neuilly-sur-Seine, a wealthy suburb of Paris, and in 2005 he became the interior minister, where he oversaw the response to the riots in the suburbs.\\
\\
Sarkozy won the presidential primary of the UMP in 2007, and went on to win the presidency in May of that year, defeating Socialist Party candidate S<C3><A9>gol<C3><A8>ne Royal in the runoff election. During his presidency, Sarkozy pursued policies on immigration, security, and economic reform. He also sought to strengthen France's relationships with other countries, particularly the United States.\\
\\
However, Sarkozy's popularity waned during his second term, and he was not renominated as the UMP's presidential candidate in 2012. He went on to lose the presidential election to Socialist Party candidate Francois Hollande. Since leaving office, Sarkozy has faced legal troubles, including being investigated for corruption and influence peddling. \\
\\
Sarkozy is married to former model and journalist Cecilia Ciganer-Albeniz, and they have three children together. He is an avid tennis player and has a reputation for being a workaholic. Despite the controversies that have dogged his political career, Sarkozy remains a prominent figure in French politics, and his influence on the center-right is still felt. \\
\midrule
Topic: \textbf{Nicolas Sarkozy}, Knowledge Source: Generated, Filtering Strategy: Internal\\
\rule{0pt}{2ex}
Nicolas Sarkozy, a French politician, was born on January 28, 1955. He was the President of France from May 16, 2007, until May 15, 2012. Sarkozy began his political career in the Rally for the Republic (RPR) party, holding various local and national positions within the party and in French politics. He served as the Minister of the Interior and a member of the French National Assembly before his political career took a steep turn when he was elected as the President of France in 2002. During his presidency, Sarkozy implemented various economic policies, including significant tax reforms, but was criticized for his controversial statements and perceived authoritarianism. He did not seek re-election and was defeated in the 2012 elections. Sarkozy continues to have a presence in French politics, despite facing numerous investigations and trials related to political financing abuses and other allegations. He also faces the possibility of legal action. \\
\bottomrule
    \end{tabular}
    \label{appendix:tab-sample-generations}
    \caption{Generated samples from two Mistral-7B-Instruct-v0.2 models. The top is generated from the model without any finetuning. The bottom is generated from a model finetuned on the Bios dataset using generated knowledge and internal filtering strategy.}
\end{table}

\end{document}